\documentclass[11pt]{article}

\usepackage[letterpaper,margin=1in]{geometry}
\usepackage{times}
\usepackage[utf8]{inputenc}
\usepackage[T1]{fontenc}
\usepackage{amsmath,amssymb,amsthm,mathrsfs}
\usepackage{graphicx,booktabs,tabularx,longtable,multirow,array,makecell}
\usepackage{microtype,xcolor,enumitem,float}
\usepackage[round,authoryear]{natbib}
\usepackage{hyperref}
\usepackage{wrapfig}
\usepackage{xurl}
\usepackage{algorithm}
\usepackage{subcaption}
\usepackage{algpseudocode}
\usepackage{fancyhdr}
\usepackage{xcolor}

\hypersetup{
  colorlinks=true,
  linkcolor=blue!45!black,
  citecolor=blue!45!black,
  urlcolor=blue!45!black,
  pdftitle={ForkPilot: Self-Evolving Policy for Retrospective Search in Long-Horizon Agents},
  pdfauthor={Xinyue Zeng, Shivam Shandilya, Guilherme Potje, Leonardo Nunes, Rakshanda Agarwal, Ranveer Chandra, Emre Kiciman, Dawei Zhou, Tusher Chakraborty}
}

\fancypagestyle{firstpageheader}{
  \fancyhf{}

  \fancyhead[L]{%
    \includegraphics[height=0.85cm]{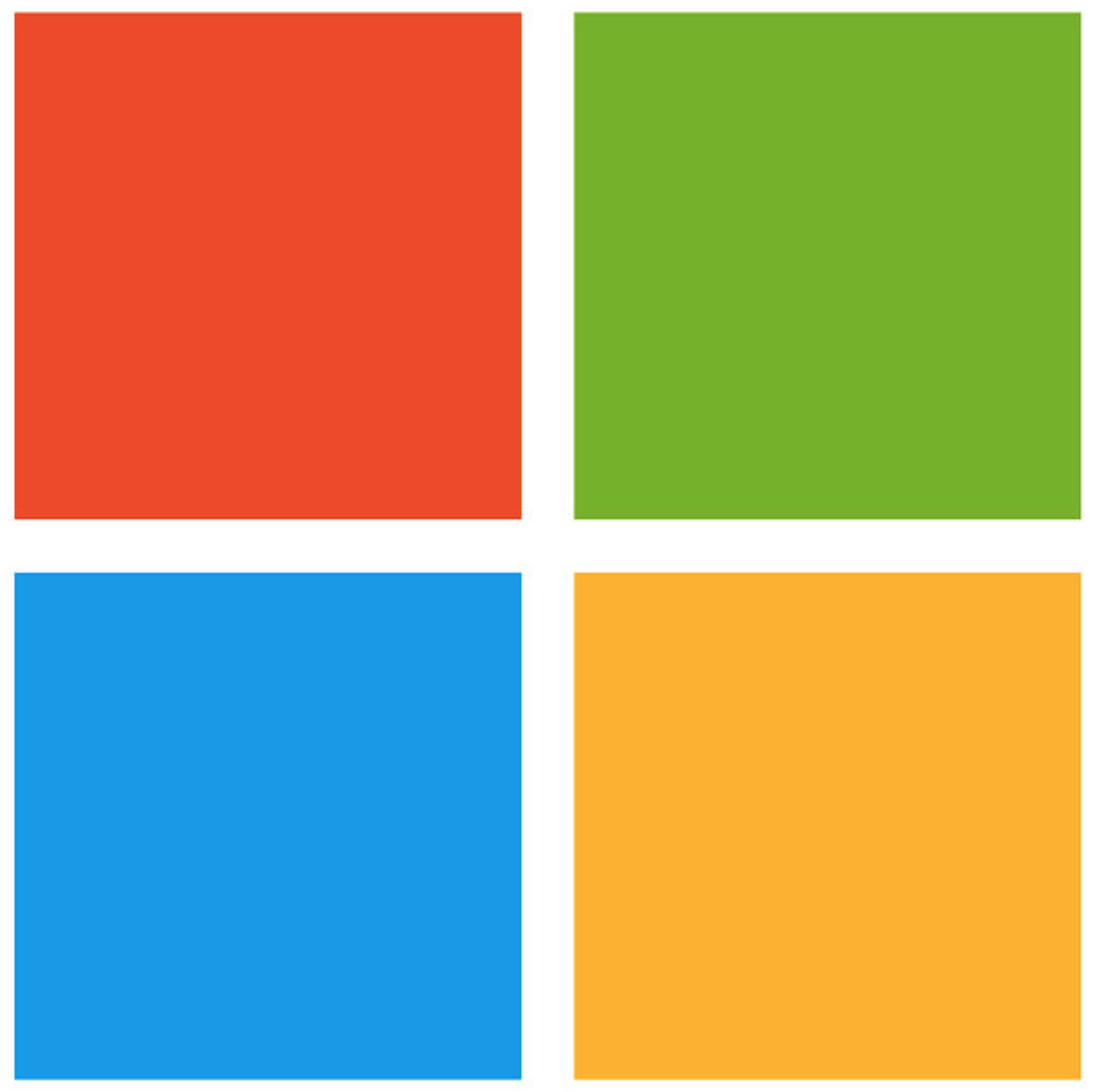}%
  }

  \fancyhead[R]{%
    \includegraphics[height=0.68cm]{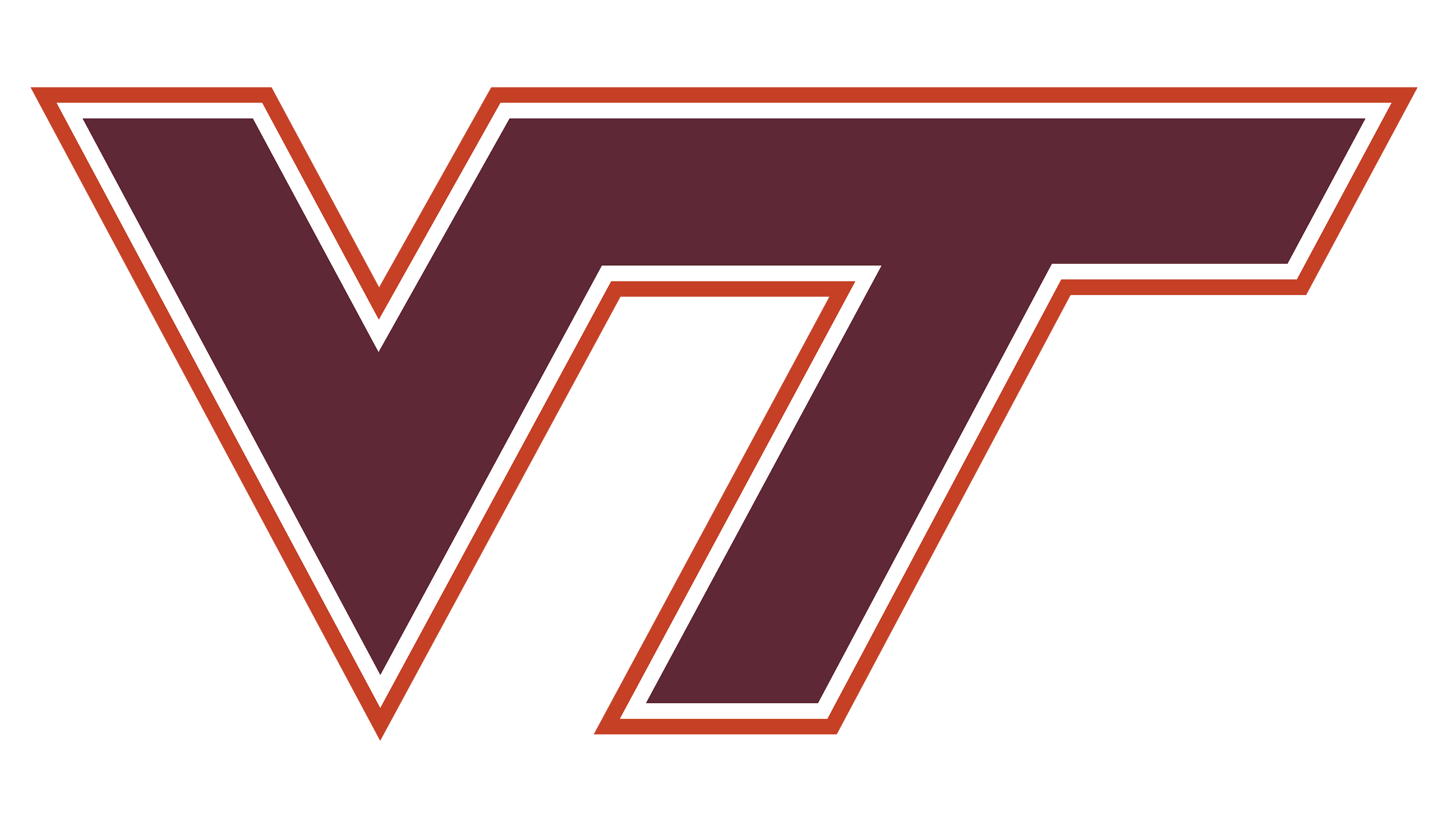}%
  }

  \renewcommand{\headrulewidth}{0.6pt}
  \renewcommand{\footrulewidth}{0pt}
}

\newcommand{\model}{\textsc{ForkPilot}}
\newcommand{\method}{\textsc{SVD}}
\newcommand{\fptaubench}{\ensuremath{\tau^3}\text{-Bench}}
\newcommand{\fpsandbox}{MM-ToolSandbox}
\newcommand{\E}{\mathbb{E}}
\newcommand{\Var}{\operatorname{Var}}
\newcommand{\KL}{D_{\mathrm{KL}}}
\newcommand{\TV}{\operatorname{TV}}

\newtheorem{problem}{Problem}
\newtheorem{theorem}{Theorem}

\setlist[itemize]{leftmargin=*,topsep=3pt,itemsep=2pt}
\setlist[enumerate]{leftmargin=*,topsep=3pt,itemsep=2pt}

\title{\model: Self-Evolving Policy for Retrospective Search in Long-Horizon Agents}
\author{Xinyue Zeng, Shivam Shandilya, Guilherme Potje, Leonardo Nunes,
Rakshanda Agarwal, Ranveer Chandra, Emre Kiciman, Dawei Zhou,
Tusher Chakraborty}
\date{}

\makeatletter
\renewcommand{\maketitle}{%

  \thispagestyle{firstpageheader}

  \begin{flushleft}

    {\fontsize{25}{29}\selectfont
     \bfseries
     \@title\par}

    \vspace{1.45em}

    {\fontsize{13}{17}\selectfont
     \bfseries
      Xinyue Zeng\textsuperscript{1,2,\textdagger},\allowbreak\ 
      Shivam Shandilya\textsuperscript{2,\textdagger},\allowbreak\ 
      Guilherme Potje\textsuperscript{2},\allowbreak\ 
      Leonardo Nunes\textsuperscript{2},\allowbreak\ 
      Rakshanda Agarwal\textsuperscript{2},\allowbreak\ 
      Ranveer Chandra\textsuperscript{2},\allowbreak\ 
      Emre Kiciman\textsuperscript{2},\allowbreak\ 
      Dawei Zhou\textsuperscript{1},\allowbreak\ 
      Tusher Chakraborty\textsuperscript{2}\par}

    \vspace{0.55em}

    {\fontsize{11}{14}\selectfont
      \textsuperscript{1}Virginia Tech
      \qquad
      \textsuperscript{2}Microsoft\par}

    \vspace{1.25em}

  \end{flushleft}

  \begingroup
    \renewcommand{\thefootnote}{\fnsymbol{footnote}}
    \footnotetext[1]{Work done during an internship at Microsoft.}
    \footnotetext[2]{Equal Contribution.}
  \endgroup
}
\makeatother

\begin{document}
\maketitle

\begin{abstract}
Interactive language-model agents increasingly solve complex tasks through long-horizon, multi-call reasoning, where errors in beliefs or actions can compound across tool interactions. Retrospective search can recover from such failures but is prone to misallocation.
Delayed outcomes obscure the contribution of intermediate search decisions, leading to \emph{Attribution Complexity}, while evolving execution evidence leads to \emph{Adaptation Complexity}, where previously learned estimates become stale. To address these challenges, we first introduce Search Value Dynamics (\method), which characterizes the evolving trade-off between the gain and cost of retrospective search. Building on \method, we propose \model{}, a self-evolving two-stage policy-learning framework. In the first stage, \model{} learns a search-value policy offline from completed trajectories through automatically constructed outcome comparisons. In the second stage, it makes search decisions on the current observations and then self-evolves, incorporating newly completed trajectories into subsequent policy updates.
We evaluate \model{} across 6 diverse benchmarks and 7 widely used LLM backbone families, including four open-source families, GPT-5.6 Sol and Opus 4.8 in a production agentic system, against 9 competitive baselines, including real-world harness deployment used by hundreds of thousands of paid users. \model{} achieves comparable state-of-the-art performance while reducing token usage by up to 59.2\%, demonstrating its efficacy.
\end{abstract}

\section{Introduction}\label{sec:introduction}
\begin{figure}[h!]
    \centering
    \vspace{-0.5em}
    \begin{minipage}[b]{0.64\linewidth}
        \centering
        \includegraphics[width=\linewidth,keepaspectratio]
        {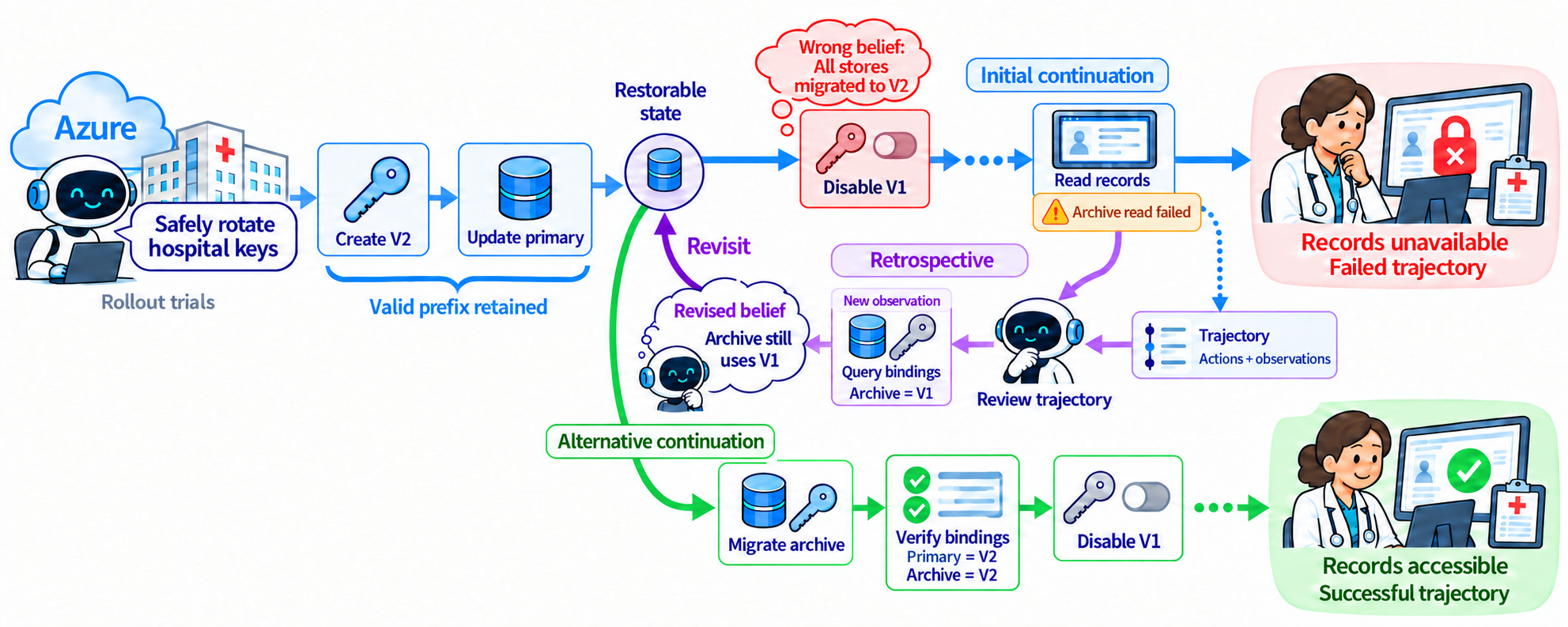}
        \par\vspace{0.25em}
        {\small\bfseries (a)}
    \end{minipage}\hfill
    \begin{minipage}[b]{0.33\linewidth}
        \centering
        \includegraphics[width=\linewidth,keepaspectratio]
        {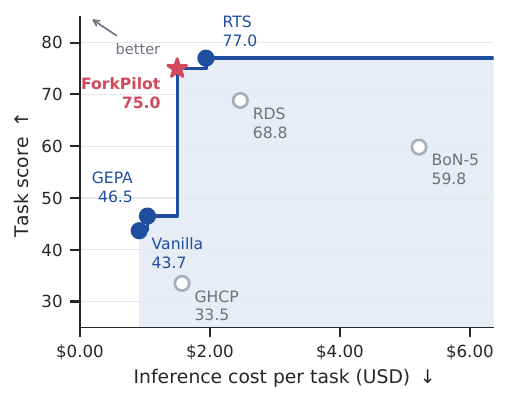}
        \par\vspace{0.25em}
        {\small\bfseries (b)}
    \end{minipage}
    \caption{
    \textbf{(a)} Retrospective search in long-horizon interaction. After an archive access failure, retrospection prompts the agent to query key bindings and revise its belief that migration is complete. The agent then explores an alternative continuation from an earlier state, preserving the valid prefix.
    \textbf{(b)} Quality-Cost Pareto frontier for GPT-5.6 Sol on AutomationBench. ForkPilot (red star) lies on the Pareto frontier, achieving comparable task quality at lower inference cost than Retrospective Tree Search (RTS), which we optimized for task reward using a heuristic selected on source validation. Vanilla denotes a single agent rollout during search. 
    }
    \label{fig:example}
    \vspace{-0.5em}
\end{figure}

Interactive language-model agents solve long-horizon tasks through tool calls, acting on evolving observations that only partially reveal the environment state~\citep{liu2024agentbench,zhou2024webarena}. Failures can stem from incorrect beliefs about this state or poor actions despite accurate beliefs~\citep{tang2026rewarding,djuhera2026tsr}. Prior work has studied information gathering through tool use~\citep{yao2023react}, reflection on task feedback~\citep{shinn2023reflexion}, and correction with external verification~\citep{gou2024critic}. Retrospective test-time search examines failed or partially successful trajectories alongside grader or verifier feedback~\citep{zhou2024lats,yu2025exact}. Actions and observations in the trajectory, together with feedback on task completion, help the agent revisit beliefs and decisions, seek further observations, and explore alternatives. When earlier states can be restored, search can fork from a state, preserving the valid action prefix while trying another continuation~\citep{koh2024tree}. In Figure~\ref{fig:example}(a), an archive access failure prompts the agent to review its trajectory and inspect key bindings. Finding that the archive still uses the old key, the agent forks from an earlier state to migrate the archive and verify the bindings before disabling the old key.
Allocating retrospective search is difficult in long-horizon tool use: observations are partial, and rewards may arrive only when execution ends~\citep{zhou2024webarena,tan2026hindsight}. Searching before enough evidence accumulates may waste forks~\citep{holt2026fact}, while waiting too long leaves less budget to recover. Different action prefixes also produce different observations and candidate actions, so another trajectory may require fewer corrections than one searched repeatedly~\citep{zhou2024lats,koh2024tree,yu2025exact}. Even an improving fork does not establish that further search on the same trajectory is worth its cost~\citep{snell2025scaling}. Its effects may emerge only after later tool calls, making its contribution difficult to separate from subsequent actions~\citep{feng2505group}. New observations can reveal recovery actions or make further search unpromising, requiring repeated decisions about when to retrospect, which trajectory to explore, and how much to search~\citep{yao2023react,zhou2024lats}. Live environments may also change independently of the agent, making learned allocation policies stale~\citep{pan2024webcanvas,xue2025illusion}.

Existing work addresses computation allocation through input-adaptive budgets, confidence-based stopping, and complexity-based token allocation~\citep{damani2025hard,fu2025certaindex,lin2026planbudget}, while learned planning policies decide when to deliberate during interaction~\citep{paglieri2025learning}. Complementarily, experience-based methods such as Reflexion~\citep{shinn2023reflexion}, ReasoningBank~\citep{ouyang2026reasoningbank}, and ACE~\citep{zhang2026ace} leverage previous trajectories and feedback to improve reasoning and guide exploration through evolving reflections, memory, and context. However, retrospective search in long-horizon, multi-call reasoning requires more than adjusting reasoning budgets or improving action selection through accumulated experience. Delayed, entangled outcomes obscure the incremental benefit of additional search and whether it justifies its computational cost, creating \textbf{Attribution Complexity}, while evolving tool observations and changes in the environment can render previously learned search-value estimates stale, creating \textbf{Adaptation Complexity}. These challenges motivate two research questions: \textbf{Q1:} \emph{How can we theoretically understand the value of retrospective search under delayed rewards and evolving observations in long-horizon, multi-turn tool interaction?} \textbf{Q2:} \emph{Can models self-evolve through accumulated experience to dynamically allocate inference resources, balancing effectiveness and efficiency?}


To answer these questions, we first introduce \textbf{Search Value Dynamics}
(\method), which characterizes the evolving trade-off between the gain and
cost of retrospective search. Building on \method, we propose
\textbf{\model{}}, a self-evolving two-stage
policy-learning framework. \model{} can bootstrap its search-value policy
from completed trajectories using automatically constructed outcome
comparisons. During deployment, it cit makes search decisions on the current observations and then self-evolves, incorporating newly completed trajectories into subsequent policy updates. This self-evolving refinement provides
a mechanism for adapting search allocation to live real-world environments
that evolve independently of the agent's actions.

We evaluate \model{} across 6 diverse benchmarks and 7 widely used LLM
backbone families. We also deploy \model{} in a production agentic system, where the agent uses GPT-5.6 Sol and Claude Opus 4.8 served through production inference platforms employing state-of-the-art caching. We compare against 9 competitive baselines, including a real-world harness deployment used by hundreds of thousands of paid users. Across all experiments, \model{} achieves comparable state-of-the-art performance while reducing token usage by up to 59.2\%, demonstrating its efficacy. As shown in Figure~\ref{fig:example}(b), \model{} lies on the empirical quality-token Pareto frontier across benchmarks and backbones.

\section{Preliminaries}\label{sec:preliminaries}


\subsection{Challenges in Long-horizon Retrospective Search}

Given task $x$ and computation budget $b$, a retrospective-search host restores an earlier environment state and forks a new branch with the black-box agent policy $\pi$ to sample alternative action sequences, preserving the action-observation prefix up to that state. At the fork, $\pi$ conditions on the preserved prefix including $x$, previously completed trajectory, and permitted feedback available at the retrospective decision~\citep{zhou2024lats,yu2025exact}.

At decision $t$, the search allocation controller observes
$z_{t}=\phi(x,e_{1:\ell_{t}},\mathcal T_{t},b_{t})$,
which summarizes observed action-observation events, explored branches and their available outcomes, and remaining budget. The controller cannot observe the full environment state or future events. From the feasible menu $\mathcal U_{t}$, it chooses an allocation $u$, such as fork count and rollout depth. $u^0$ signifies that the rollout execution continues without any forking from the current state.

We define the value of $u$ relative to $u^0$ using paired replay $\Gamma$. Both actions start from the same restored state under coordinated randomness, a common subsequent execution rule, and the same total-budget constraint (Appendix~\ref{app:paired_protocol}). Let $\tau^u_{t}$ and $\tau^0_{t}$ denote the completed continuations and their search records, including any later forks and branch rollouts. Then
$
Y_{t}(u)
=R(x,\tau^u_{t})-R(x,\tau^0_{t}),
$
$
\Delta C_{t}(u)
=C(\tau^u_{t})-C(\tau^0_{t}),
$
$
G_{t}(u)
=\E_\Gamma\!\left[Y_{t}(u)-\lambda\Delta C_{t}(u)\mid z_{t}\right],
G_{t}(u^0)=0,
$
where $\lambda\ge0$ and $C$ includes the generation and evaluation of unselected branches. Completed pairs provide learning and evaluation targets; deployment does not require paired replay at every decision.

\textbf{Attribution Complexity.}
A completed reward reflects both the current fork and later tool actions. Sampling and environment responses therefore make $Y_{t}(u)$ variable even at a fixed $z_{t}$. We quantify this variability as
$\mathcal A^\Gamma_{t}(u)=\Var_\Gamma[Y_{t}(u)\mid z_{t}]$.
For $n$ conditionally independent paired replays at fixed $(z_{t},u)$, the sample-mean variance is $\mathcal A^\Gamma_{t}(u)/n$. Larger attribution complexity thus requires more completed pairs for the same sampling precision.

\textbf{Adaptation Complexity.}
New observations and search outcomes change $z_{t}$; changes in the environment may also alter reward effects at a fixed input. Let $P$ and $Q$ be the current and source conditional laws of $Y_{t}(u)$ at the same $z_{t}$, defined under a common $\Gamma$ and subsequent execution rule. 

Since $R\in[0,1]$ implies $Y_{t}(u)\in[-1,1]$,
$
\left|\E_PY_{t}(u)-\E_QY_{t}(u)\right|
\le \mathcal S_{t}(u)
:=2\TV(P,Q)
\le\sqrt{2\KL(P\Vert Q)},
$
where the last inequality assumes finite KL divergence and natural logarithms. This bound describes source-to-current shift at a fixed input. 

\subsection{Problem Definition}\label{sec:problem}
The host admits only feasible actions and reserves resources for subsequent execution, enforcing $C(\tau)\le B$ for each complete task execution $\tau$. We select $\lambda$ using source validation only: $B$ imposes the hard constraint, while $\lambda$ governs the reward-cost trade-off among feasible decisions. 
Self-evolution proceeds across completed source tasks. The controller uses
$\theta_n$, learned from $\mathcal D_n$, throughout the next task. Only after
that task finishes does it incorporate its completed experience
$\xi_{n+1}$ and update:
$
\mathcal D_{n+1}=\mathcal D_n\cup\{\xi_{n+1}\},
\theta_{n+1}=\mathsf{Update}(\theta_n,\mathcal D_{n+1}).
$
Here, $\xi_{n+1}$ contains the observed search decisions, continuations, outcomes, and costs; the update does not require labels specifying optimal search allocations.
 
\begin{problem}[Efficient Self-evolving Policy on Retrospective Long-horizon Reasoning]
\label{prob:budgeted_control}
\mbox{}\\
\noindent
\textbf{Given.} A policy $\pi$, a retrospective-search host, completed experience $\mathcal D_n$ from $n$ source tasks, and, at each eligible decision, the observable input $z_t$ and feasible menu $\mathcal U_t$.
 
\textbf{Find.} A learning procedure and its induced sequence of causal policies $\{\mu_{\theta_n}\}_{n\geq 0}$, where each policy is fitted from permitted completed source experience and selects $\mu_{\theta_n}(z_t,\mathcal U_t)\in\mathcal U_t$ using only information available at that decision.  At each stage, the policy approximates the constrained objective
\begin{equation}
\max_{\mu\in\Pi_{\mathrm{causal}}}\quad
\E\!\left[R(x,\tau)-\lambda C(\tau)\right]
\quad \text{s.t.} C(\tau)\le B\quad\text{a.s.}
\label{eq:control_objective}
\end{equation}
Here, $\Pi_{\mathrm{causal}}$ is the class of policies that select actions from the current observable input and feasible menu, using parameters fitted only on permitted completed experience.
\end{problem}

\section{Learning Retrospective Search Allocation}\label{sec:svd}
Building on the problem formulation, we next characterize the uncertainty in learning paired effects and the changes incurred when using a source estimate at a later decision. These motivate the two stages of \model{}.

\subsection{Search-value dynamics}\label{sec:svd_analysis}

Following classical metareasoning~\citep{russell1991metareasoning,hay2012selecting}, fix a source-task cost scale $\bar C_{\mathrm{src}}>0$. Suppressing decision subscripts when unambiguous, write $D(u)=\Delta C(u)/\bar C_{\mathrm{src}}$ and $\lambda_D=\lambda\bar C_{\mathrm{src}}$. The paired net effect is $W(u)=Y(u)-\lambda_DD(u)=Y(u)-\lambda\Delta C(u)$. Let $z_{\mathrm{src}}$ denote an input represented in completed source experience and $z_t$ the current input. We use $\mathsf Q$ for the source conditional law of $(Y,D)$ and $\mathsf P$ for the corresponding current law. Both laws use the same fixed $\Gamma$, including $\rho$; the comparison does not change the continuation rule while changing the input or task law.
 
\textbf{Attribution complexity.} The first challenge is estimating whether an allocation is worth its incremental cost. At $(z_{\mathrm{src}},u)$, let $A_{\mathrm{src}}^\Gamma(u)=\Var_{\mathsf Q}[Y(u)\mid z_{\mathrm{src}}]$ and $V_{\mathrm{src}}^\Gamma(u)=\Var_{\mathsf Q}[W(u)\mid z_{\mathrm{src}}]$. Then
$
V_{\mathrm{src}}^\Gamma(u)
=
A_{\mathrm{src}}^\Gamma(u)
+\lambda_D^2
\Var_{\mathsf Q}[D(u)\mid z_{\mathrm{src}}]
-2\lambda_D
\operatorname{Cov}_{\mathsf Q}
\!\left(Y(u),D(u)\mid z_{\mathrm{src}}\right).
$

Thus both reward and cost variation matter. With $n$ conditionally independent pairs at the same input and action, the variance of their mean net effect is $V_{\mathrm{src}}^\Gamma(u)/n$.
 
The controller representation can add another source of ambiguity. At a source decision, let $H$ be the available history, $Z=\phi(H)$ its representation, and $g(H,u)=\E_{\mathsf Q}[W(u)\mid H]$ for an action feasible across the histories considered. Because $Z$ is determined by $H$,
\begin{equation}
\begin{aligned}
\Var_{\mathsf Q}(W(u)\mid Z)
={}&\E_{\mathsf Q}[\Var_{\mathsf Q}(W(u)\mid H)\mid Z] +\Var_{\mathsf Q}(g(H,u)\mid Z).
\end{aligned}
\label{eq:representation_attribution}
\end{equation}
The second term arises when different permitted histories with different search values map to the same input $Z$. Sampling noise and this representation ambiguity jointly complicate attribution.
 
\textbf{Adaptation complexity.} A source estimate may become stale when observations, explored branches, or the remaining budget change the controller input. The feasible menu can change as well. These changes motivate recomputing action scores at each eligible decision. They do not, by themselves, show that a newly fitted prediction is more accurate; the following bound only characterizes the discrepancy incurred by reusing a source-input estimate.
 
We can also bound the discrepancy between a source estimate and the current value. For $\mathsf L\in\{\mathsf P,\mathsf Q\}$, define $m_{\mathsf L}(z,u)=\E_{\mathsf L}[Y(u)\mid z]$ and $c_{\mathsf L}(z,u)=\E_{\mathsf L}[D(u)\mid z]$. For an action feasible at both $z_{\mathrm{src}}$ and $z_t$, define
\begin{align*}
\Omega_{\mathrm{src}\to t}(u)
&=|m_{\mathsf P}(z_t,u)-m_{\mathsf P}(z_{\mathrm{src}},u)|,\\
S_{\mathrm{src}}(u)
&=2\TV(\mathsf P^u_{z_{\mathrm{src}}},\mathsf Q^u_{z_{\mathrm{src}}}),\\
\xi_{\mathrm{src},t}(u)
&=|c_{\mathsf P}(z_t,u)-c_{\mathsf Q}(z_{\mathrm{src}},u)|,
\end{align*}
where $\mathsf P^u_z$ and $\mathsf Q^u_z$ are the conditional laws of $Y(u)$ at $z$. The terms measure input change under the current law, source-to-current reward-law shift at the source input, and the combined change in expected normalized cost. Because $Y(u)\in[-1,1]$ when $R\in[0,1]$, $S_{\mathrm{src}}(u)\le\sqrt{2\KL(\mathsf P^u_{z_{\mathrm{src}}}\Vert\mathsf Q^u_{z_{\mathrm{src}}})}$ whenever this KL divergence is finite.
 
\begin{theorem}[Source-to-current search-value bound]
\label{thm:search_value_dynamics}
Let $\mathcal F_t=\mathcal U_t\setminus\{u^0\}$ contain $K\ge1$ actions, each feasible at $z_{\mathrm{src}}$. Fix $\Gamma$, the source and current laws, $z_{\mathrm{src}}$, $z_t$, $\mathcal F_t$, and the sample counts $n_u\ge1$ independently of the replay draws used below. Suppose $\mathsf P^u_{z_{\mathrm{src}}}$ and $\mathsf Q^u_{z_{\mathrm{src}}}$ are well defined for every $u\in\mathcal F_t$ (i.e.\ $z_{\mathrm{src}}$ has positive probability or density under the respective marginal law of $Z$), with the analogous current-law quantities well defined at $z_t$. For each $u\in\mathcal F_t$, draw $n_u$ conditionally i.i.d.\ source pairs at $(z_{\mathrm{src}},u)$ under $\mathsf Q$, with finite $V_{\mathrm{src}}^\Gamma(u)$. Let $\overline W_{\mathrm{src}}(u)$ be their mean, $\widehat G_{\mathrm{src}}(u)$ any source-fitted predictor evaluated at $z_{\mathrm{src}}$, and $e_{\mathrm{src}}(u)=|\widehat G_{\mathrm{src}}(u)-\overline W_{\mathrm{src}}(u)|$. If rewards lie in $[0,1]$, then for $0<\delta<1$, with probability at least $1-\delta$ over these source replay draws, simultaneously for all $u\in\mathcal F_t$,
\begin{equation}
\begin{aligned}
|G_t(u)-\widehat G_{\mathrm{src}}(u)|
&\le e_{\mathrm{src}}(u)
+\sqrt{\frac{K V_{\mathrm{src}}^\Gamma(u)}{n_u\delta}}\\
&\quad+\Omega_{\mathrm{src}\to t}(u)+S_{\mathrm{src}}(u)\\
&\quad+\lambda_D\xi_{\mathrm{src},t}(u)
=:\beta_{\mathrm{src},t}^\delta(u),
\end{aligned}
\label{eq:svd_decomp}
\end{equation}
where $G_t(u)=m_{\mathsf P}(z_t,u)-\lambda_Dc_{\mathsf P}(z_t,u)$.
\end{theorem}
 
Additional pairs reduce the displayed sampling term. The condition $n_u\ge K V_{\mathrm{src}}^\Gamma(u)/(\delta\eta^2)$ is sufficient to make that term at most $\eta$; it does not remove input or law changes. Set $\widehat G_{\mathrm{src}}(u^0)=\beta_{\mathrm{src},t}^\delta(u^0)=0$ and define $L_t^\delta(u)=\widehat G_{\mathrm{src}}(u)-\beta_{\mathrm{src},t}^\delta(u)$ and $U_t^\delta(u)=\widehat G_{\mathrm{src}}(u)+\beta_{\mathrm{src},t}^\delta(u)$. On the theorem event, if $u_L\in\arg\max_{u\in\mathcal F_t}L_t^\delta(u)$ has $L_t^\delta(u_L)>0$, then $G_t(u_L)>0$ and its local net-value regret relative to $u^*\in\arg\max_{u\in\mathcal U_t}G_t(u)$ is at most $2\beta_{\mathrm{src},t}^\delta(u^*)$. If $\max_{u\in\mathcal F_t}U_t^\delta(u)\le0$, skip is optimal. Appendix~\ref{app:proofs} gives the proof.
 
The bound concerns reusing $\widehat G_{\mathrm{src}}(u)$ at a fixed current input. Its probability statement is not uniform over inputs or menus chosen adaptively after observing the replay draws. At runtime, the learned predictor is instead evaluated on $z_t$; neither this bound nor $\beta_{\mathrm{src},t}^\delta(u)$ controls that refreshed prediction's error. The exact-input pairs and shift terms are also unavailable during deployment. Thus the theorem diagnoses why attribution and stale estimates matter under a fixed replay rule; it provides no certificate for the learned margin or the deployed controller.

\subsection{\model{}: Two-stage Self-evolving Policy}\label{sec:forkpilot}
\model{} implements these requirements in two stages. Stage I learns search effects and their reliability from paired completed trajectories, which may be collected both during an offline initialization phase and from tasks subsequently completed during deployment. These paired replays follow the same reference rule $\rho$ and form a source pool $\mathcal D_n$. As additional deployment tasks are completed, newly available trajectory pairs extend $\mathcal D_n$ to $\mathcal D_{n'}$, where $n'>n$, and Stage I refits the controller parameters $\theta_{n'}$ on the expanded pool before a subsequent evaluation stage. Stage II operates within each evaluated task: the fitted controller parameters remain fixed, while search allocations are repeatedly recomputed as new trajectory evidence becomes available. Thus, Stage I enables the policy to evolve across completed tasks, whereas Stage II adapts its search decisions within a task without online parameter updates.
 
\textbf{Stage I: Attribution from completed experience.} At an eligible source state, paired replay compares $u$ with $u^0$ from the same restorable state. The completed executions yield $\widetilde Y=R(x,\tau^u)-R(x,\tau^0)$ and $\widetilde D=[C(\tau^u)-C(\tau^0)]/\bar C_{\mathrm{src}}$. Each pair labels its selected action only. Training needs neither human allocation labels nor paired replay at every deployment decision.
 
An encoder produces $r_t=f_\psi(z_t)$. An action-conditioned ensemble of $M$ task-bootstrap heads predicts reward effects $\widehat m_t^{(j)}(u)$ and normalized cost effects $\widehat c_t^{(j)}(u)$. Averaging across heads gives $\widehat G_t(u)=\widehat m_t(u)-\lambda_D\widehat c_t(u)$, with the fitted ensemble evaluated on the current $z_t$ at deployment. Because the theorem's shift terms are unavailable at runtime, \model{} estimates an empirical margin $\widehat\beta_t(u)$ using ensemble disagreement, source-representation support, and held-out paired residuals. A margin model predicts an upper quantile of $|\widetilde Y-\lambda_D\widetilde D-\widehat G_t(u)|$, with a correction calibrated on separate source data (Appendix~\ref{app:controller}). We set $L_t(u)=\widehat G_t(u)-\widehat\beta_t(u)$ and $U_t(u)=\widehat G_t(u)+\widehat\beta_t(u)$. These are decision scores, not conditional-coverage certificates or estimates of the theorem's unavailable radius.
 
\textbf{Stage II: Closed-loop adaptation.} 
With the fitted artifact $\theta_{n'}$ fixed, the controller scores the current feasible menu using $z_t$. Let $\mathcal F_t=\mathcal U_t\setminus\{u^0\}$, and let $k(u)$ and $d(u)$ denote an allocation's fork count and rollout-depth limit. Fix probe-specific caps $k_{\mathrm{pr}}$ and $d_{\mathrm{pr}}$ before evaluation, stricter than the general host limits, and set $\mathcal P_t=\{u\in\mathcal F_t:k(u)\le k_{\mathrm{pr}},\ d(u)\le d_{\mathrm{pr}}\}$ (Appendix~\ref{app:controller}). Define $L_t^{\max}=\max_{u\in\mathcal F_t}L_t(u)$ and $U_t^{\mathrm{probe}}=\max_{u\in\mathcal P_t}U_t(u)$, taking a maximum over an empty set to be $-\infty$. The controller chooses
\begin{equation}
u_t=
\begin{cases}
\arg\max_{u\in\mathcal F_t}L_t(u),
&L_t^{\max}>0,\\
\arg\min_{\substack{u\in\mathcal P_t\\U_t(u)>0}}\widehat c_t(u),
&L_t^{\max}\le0<U_t^{\mathrm{probe}},\\
u^0,&\text{otherwise}.
\end{cases}
\label{eq:runtime_rule}
\end{equation}
Thus \model{} selects a search action with positive lower score, chooses a capped probe with the lowest predicted normalized full-execution cost difference when an upper score is positive, and otherwise skips the current round. 

\section{Experiments}\label{sec:experiments}
We comprehensively evaluate \model{} against 9 competitive baselines across 6 diverse benchmarks on 7 backbone families. Our experiments answer the following four questions: \textbf{\emph{Q1:}} \emph{Does \model{} consistently achieve efficiency across diverse benchmarks?}
\textbf{\emph{Q2:}} \emph{Does \model{} consistently consistently achieve efficiency across different backbone model scales?}
\textbf{\emph{Q3:}} \emph{Does \model{} retain its efficiency gains on frontier black-box models?}
\textbf{\emph{Q4:}} \emph{Does \model{} require privileged feedback or large bootstrap samples?}

\subsection{Experimental Settings}
\label{sec:experiment_setup}

\textbf{Baselines.}
We compare against 9 competitive baselines, including three families of methods: (1) \emph{test-time search}, including BoN-5~\citep{snell2025scaling}, full Retrospective Tree Search (RTS)
and Retrospective Depth Search (RDS), where RTS uses a source-validation-selected heuristic to explore alternative continuations from earlier states for strong task reward and RDS reviews one initial rollout and continues only from its final resumable node; (2) \emph{experience-based adaptation}, including ACE~\citep{zhang2026ace}, ReasoningBank~\citep{ouyang2026reasoningbank}, Reflexion~\citep{shinn2023reflexion}, GEPA~\citep{agrawal2026gepa}, and Trace2Skill~\citep{ni2026trace2skill}; and (3) \emph{state-of-the-art agent harnesses}, represented by GitHub Copilot (GHCP)~\footnote{\url{https://code.visualstudio.com/blogs/2026/05/15/agent-harnesses-github-copilot-vscode}}, which repeatedly conditions action selection on accumulated interaction history and tool results without explicit retrospective trajectory search. 

\begin{table}[h!]
\centering
\caption{Efficiency comparison across 6 diverse benchmarks. $S$ is task score (0-100), Higher $S$ is better; $R/U$ are raw/uncached token ratios to RTS. Lower $R/U$ is better. The best is in \textbf{Bold} and the second best is in \underline{underline}.}
\label{tab:fp_main}
\begingroup
\scriptsize
\setlength{\tabcolsep}{3.6pt}
\begin{tabular}{@{}l*{18}{r}@{}}
\toprule
Method & \multicolumn{3}{c}{Automation} & \multicolumn{3}{c}{$\tau^3$-Bench} & \multicolumn{3}{c}{SWE-bench} & \multicolumn{3}{c}{MM-ToolSandbox} & \multicolumn{3}{c}{WorkArena++} & \multicolumn{3}{c}{VitaBench}\\
\cmidrule(lr){2-4}\cmidrule(lr){5-7}\cmidrule(lr){8-10}\cmidrule(lr){11-13}\cmidrule(lr){14-16}\cmidrule(lr){17-19}
& $S$ & $R$ & $U$ & $S$ & $R$ & $U$ & $S$ & $R$ & $U$ & $S$ & $R$ & $U$ & $S$ & $R$ & $U$ & $S$ & $R$ & $U$\\
\midrule
\multicolumn{19}{l}{\textbf{DeepSeek-V4-Flash}}\\
BoN-5 & 49.5 & 1.17 & 1.26 & 74.1 & 1.18 & 1.24 & 72.0 & 1.10 & 1.22 & 28.2 & 1.07 & 1.26 & 49.4 & 1.08 & 1.29 & 28.6 & 1.08 & 1.29\\
RTS & 56.8 & 1.00 & 1.00 & 82.1 & 1.00 & 1.00 & 78.8 & 1.00 & 1.00 & 35.2 & 1.00 & 1.00 & \underline{52.1} & 1.00 & 1.00 & \textbf{32.6} & 1.00 & 1.00\\
ACE & 55.0 & 0.62 & 0.66 & 81.2 & 0.69 & 0.74 & 76.7 & \underline{0.64} & \underline{0.72} & 33.1 & \underline{0.65} & \underline{0.73} & 51.9 & \textbf{0.61} & \textbf{0.67} & 31.5 & 0.60 & 0.67\\
ReasoningBank & 55.0 & 0.61 & 0.67 & \textbf{82.6} & \textbf{0.64} & 0.73 & 78.6 & \textbf{0.63} & \textbf{0.70} & \underline{35.4} & \underline{0.65} & 0.74 & 51.6 & \underline{0.62} & 0.70 & 32.0 & \underline{0.59} & \textbf{0.63}\\
Reflexion & 53.8 & 0.72 & 0.79 & 78.0 & 0.76 & 0.87 & 76.4 & 0.84 & 0.89 & 32.2 & 0.79 & 0.88 & 49.9 & 0.74 & 0.80 & 31.1 & 0.81 & 0.92\\
GEPA & \underline{57.0} & \underline{0.59} & \underline{0.65} & 80.6 & 0.70 & 0.77 & \textbf{79.5} & 0.68 & \underline{0.72} & \textbf{35.7} & 0.70 & 0.80 & 50.5 & 0.64 & 0.70 & 31.5 & \textbf{0.58} & \underline{0.64}\\
Trace2Skill & 55.8 & 0.63 & 0.71 & \underline{82.4} & 0.66 & \underline{0.72} & \underline{79.4} & 0.68 & 0.74 & 34.4 & \textbf{0.63} & \textbf{0.71} & 51.3 & 0.66 & 0.71 & 32.1 & 0.65 & 0.70\\
\model{} & \textbf{57.2} & \textbf{0.56} & \textbf{0.64} & 82.3 & \underline{0.65} & \textbf{0.69} & 78.7 & 0.69 & 0.73 & \textbf{35.7} & 0.66 & \textbf{0.71} & \textbf{52.4} & 0.64 & \underline{0.68} & \underline{32.4} & 0.65 & 0.71\\
\midrule
\multicolumn{19}{l}{\textbf{Kimi-K3}}\\
BoN-5 & 54.0 & 1.10 & 1.19 & 82.5 & 1.16 & 1.25 & 76.6 & 1.13 & 1.21 & 37.2 & 1.14 & 1.24 & 51.5 & 1.15 & 1.23 & 28.8 & 1.19 & 1.30\\
RTS & \textbf{64.5} & 1.00 & 1.00 & 86.3 & 1.00 & 1.00 & \underline{81.4} & 1.00 & 1.00 & \textbf{43.4} & 1.00 & 1.00 & 59.5 & 1.00 & 1.00 & \underline{36.3} & 1.00 & 1.00\\
ACE & 61.2 & \underline{0.59} & \underline{0.71} & 86.5 & 0.67 & \textbf{0.67} & 78.0 & \textbf{0.62} & \textbf{0.73} & 41.1 & 0.66 & 0.79 & \underline{59.8} & \underline{0.56} & 0.70 & 34.0 & \underline{0.62} & 0.76\\
ReasoningBank & 60.3 & 0.60 & 0.73 & 85.5 & 0.56 & \underline{0.70} & 81.0 & 0.68 & 0.77 & 40.3 & 0.65 & 0.78 & 57.2 & 0.62 & 0.75 & 33.5 & 0.63 & 0.77\\
Reflexion & 58.0 & 0.78 & 0.85 & 83.2 & 0.76 & 0.84 & 78.8 & 0.83 & 0.89 & 39.1 & 0.79 & 0.88 & 54.8 & 0.81 & 0.89 & 31.5 & 0.84 & 0.93\\
GEPA & 62.8 & 0.62 & 0.72 & \underline{87.0} & \underline{0.55} & \textbf{0.67} & 80.0 & 0.67 & 0.78 & 41.5 & \underline{0.64} & \underline{0.75} & 59.0 & 0.58 & \underline{0.69} & 34.8 & \underline{0.62} & \underline{0.73}\\
Trace2Skill & 62.3 & \textbf{0.57} & \underline{0.71} & 85.9 & 0.58 & \underline{0.70} & 79.6 & \underline{0.64} & 0.77 & 42.6 & \textbf{0.63} & 0.76 & \textbf{60.5} & \textbf{0.54} & \textbf{0.67} & 35.5 & \textbf{0.60} & \textbf{0.72}\\
\model{} & \underline{64.2} & 0.63 & \textbf{0.67} & \textbf{87.2} & \textbf{0.53} & 0.72 & \textbf{81.6} & 0.70 & \underline{0.74} & \underline{43.0} & 0.68 & \textbf{0.74} & \underline{59.8} & 0.64 & 0.70 & \textbf{36.5} & 0.66 & \textbf{0.72}\\
\midrule
\multicolumn{19}{l}{\textbf{DeepSeek-R1-671B}}\\
BoN-5 & 34.1 & 1.14 & 1.22 & 62.8 & 1.16 & 1.22 & 55.5 & 1.12 & 1.21 & 17.4 & 1.09 & 1.23 & 34.4 & 1.15 & 1.27 & 12.9 & 1.08 & 1.29\\
RTS & \underline{37.6} & 1.00 & 1.00 & 66.2 & 1.00 & 1.00 & \underline{62.5} & 1.00 & 1.00 & 20.9 & 1.00 & 1.00 & 40.3 & 1.00 & 1.00 & 18.2 & 1.00 & 1.00\\
ACE & 35.6 & 0.69 & 0.77 & 66.2 & \underline{0.64} & 0.75 & 60.8 & 0.69 & 0.79 & 19.9 & \underline{0.68} & \underline{0.75} & 40.0 & \underline{0.63} & 0.72 & \textbf{18.4} & 0.68 & 0.76\\
ReasoningBank & 36.4 & \underline{0.67} & 0.74 & 65.3 & 0.70 & 0.75 & 61.4 & 0.74 & 0.80 & 19.2 & 0.73 & 0.78 & 39.0 & 0.65 & 0.72 & 16.9 & 0.69 & 0.75\\
Reflexion & 33.2 & 0.77 & 0.82 & 62.8 & 0.81 & 0.90 & 61.2 & 0.81 & 0.88 & 16.4 & 0.81 & 0.87 & 37.4 & 0.76 & 0.83 & 15.7 & 0.80 & 0.89\\
GEPA & 36.4 & \textbf{0.64} & \underline{0.73} & 64.6 & 0.67 & \underline{0.72} & \underline{62.5} & \underline{0.67} & 0.77 & \textbf{21.5} & \textbf{0.63} & \textbf{0.73} & 39.2 & 0.67 & 0.77 & 17.9 & \textbf{0.64} & \textbf{0.68}\\
Trace2Skill & 36.7 & 0.71 & 0.80 & \underline{66.4} & \textbf{0.60} & \textbf{0.66} & \textbf{62.8} & \textbf{0.65} & \textbf{0.72} & 19.8 & \underline{0.68} & 0.77 & \underline{40.5} & \textbf{0.59} & \textbf{0.65} & \underline{18.3} & 0.73 & 0.79\\
\model{} & \textbf{38.0} & \textbf{0.64} & \textbf{0.70} & \textbf{66.5} & 0.66 & \underline{0.72} & 62.2 & 0.69 & \underline{0.73} & \underline{21.3} & \underline{0.68} & \textbf{0.73} & \textbf{40.7} & 0.65 & \underline{0.71} & \textbf{18.4} & \underline{0.66} & \underline{0.71}\\
\midrule
\multicolumn{19}{l}{\textbf{Qwen3.8-27B}}\\
BoN-5 & 18.0 & 1.07 & 1.16 & 66.4 & 1.13 & 1.22 & 49.2 & 1.10 & 1.18 & 15.9 & 1.11 & 1.21 & 29.0 & 1.12 & 1.20 & 9.3 & 1.16 & 1.27\\
RTS & \underline{30.5} & 1.00 & 1.00 & 73.8 & 1.00 & 1.00 & \underline{57.6} & 1.00 & 1.00 & \underline{23.3} & 1.00 & 1.00 & 38.7 & 1.00 & 1.00 & \textbf{17.5} & 1.00 & 1.00\\
ACE & 27.8 & \underline{0.56} & \underline{0.68} & 74.2 & \textbf{0.50} & \textbf{0.64} & 53.8 & \textbf{0.59} & \textbf{0.70} & 20.5 & 0.63 & 0.76 & 39.0 & \underline{0.53} & 0.67 & 15.5 & \underline{0.59} & 0.73\\
ReasoningBank & 26.8 & 0.57 & 0.70 & 73.0 & 0.53 & 0.67 & 57.0 & 0.65 & \underline{0.74} & 19.8 & 0.62 & 0.75 & 36.5 & 0.59 & 0.72 & 14.5 & 0.60 & 0.74\\
Reflexion & 23.5 & 0.75 & 0.82 & 68.9 & 0.73 & 0.81 & 52.6 & 0.80 & 0.86 & 17.8 & 0.76 & 0.85 & 33.2 & 0.78 & 0.86 & 12.5 & 0.81 & 0.90\\
GEPA & 29.0 & 0.59 & 0.69 & \textbf{75.1} & \underline{0.52} & \textbf{0.64} & 56.0 & 0.64 & 0.75 & 21.3 & \underline{0.61} & \underline{0.72} & 38.3 & 0.55 & 0.66 & 16.5 & \underline{0.59} & 0.70\\
Trace2Skill & 28.2 & \textbf{0.54} & \underline{0.68} & 73.4 & 0.55 & 0.67 & 55.2 & \underline{0.61} & \underline{0.74} & 22.1 & \textbf{0.60} & 0.73 & \underline{39.1} & \textbf{0.51} & \textbf{0.64} & 17.0 & \textbf{0.57} & \underline{0.69}\\
\model{} & \textbf{31.0} & 0.58 & \textbf{0.63} & \underline{74.4} & 0.61 & \underline{0.66} & \textbf{58.0} & 0.65 & \textbf{0.70} & \textbf{23.6} & \textbf{0.60} & \textbf{0.66} & \textbf{39.3} & 0.59 & \underline{0.65} & \underline{17.3} & 0.62 & \textbf{0.68}\\
\bottomrule
\end{tabular}
\endgroup
\end{table}

\textbf{Backbones.}
We evaluate on 7 backbone families, including both (1) \emph{open-weight models}, including DeepSeek-V4-Flash~\citep{fp_v4flash}, Kimi-K3~\citep{fp_kimik3}, DeepSeek-R1-671B~\citep{fp_r1}, Qwen3.5(4B, 9B, 28B, 122B)~\citep{fp_qwen359} and Qwen3.8-27B~\citep{fp_qwen3827} and (2) \emph{black-box models}, including GPT-5.6 Sol\footnote{\url{https://deploymentsafety.openai.com/gpt-5-6}} and Opus 4.8\footnote{\url{https://www.anthropic.com/news/claude-opus-4-8}} with \textbf{extra-high} reasoning effort.

\textbf{Benchmarks.}
We evaluate on 6 diverse benchmarks, covering workflow automation (AutomationBench~\citep{fp_automationbench}), tool-user interaction (\fptaubench{}~\citep{fp_taubench}), software engineering (SWE-bench Verified~\citep{fp_swebench}), multimodal tool use (\fpsandbox{}~\citep{fp_mmtoolsandbox}), web interaction (WorkArena++~\citep{fp_workarena}) and general interactive tasks (VitaBench~\citep{fp_vitabench}). Additional details are in Appendix~\ref{app:controller}-\ref{app:protocol}.

\begin{table}[h!]
\centering
\caption{Efficiency comparison across different backbone scales on six benchmarks. $S$ is task score (0-100); $R/U$ are raw/uncached token ratios to RTS. Lower $R/U$ is better. The best is in \textbf{Bold} and the second best is in \underline{underline}.}
\label{tab:fp_main2}
\begingroup
\scriptsize
\setlength{\tabcolsep}{3.6pt}
\begin{tabular}{@{}l*{18}{r}@{}}
\toprule
Method & \multicolumn{3}{c}{Automation} & \multicolumn{3}{c}{$\tau^3$-Bench} & \multicolumn{3}{c}{SWE-bench} & \multicolumn{3}{c}{MM-ToolSandbox} & \multicolumn{3}{c}{WorkArena++} & \multicolumn{3}{c}{VitaBench}\\
\cmidrule(lr){2-4}\cmidrule(lr){5-7}\cmidrule(lr){8-10}\cmidrule(lr){11-13}\cmidrule(lr){14-16}\cmidrule(lr){17-19}
& $S$ & $R$ & $U$ & $S$ & $R$ & $U$ & $S$ & $R$ & $U$ & $S$ & $R$ & $U$ & $S$ & $R$ & $U$ & $S$ & $R$ & $U$\\
\midrule
\multicolumn{19}{l}{\textbf{Qwen3.5-4B}}\\
BoN-5 & 9.5 & 1.13 & 1.22 & 41.2 & 1.07 & 1.28 & 18.0 & 1.18 & 1.28 & 6.0 & 1.10 & 1.28 & 13.1 & 1.12 & 1.23 & 6.5 & 1.17 & 1.22\\
RTS & \underline{14.1} & 1.00 & 1.00 & 43.6 & 1.00 & 1.00 & 24.8 & 1.00 & 1.00 & 12.9 & 1.00 & 1.00 & 20.3 & 1.00 & 1.00 & \underline{8.8} & 1.00 & 1.00\\
ACE & 13.5 & 0.71 & 0.79 & 42.2 & 0.74 & 0.83 & 24.9 & 0.77 & 0.83 & 11.8 & \underline{0.65} & \underline{0.72} & 19.2 & 0.71 & 0.81 & 6.8 & 0.77 & 0.85\\
ReasoningBank & 12.5 & \textbf{0.69} & 0.78 & 43.9 & 0.74 & 0.83 & 23.2 & 0.74 & 0.83 & 10.6 & 0.67 & 0.78 & 19.8 & 0.77 & 0.87 & 8.5 & \underline{0.72} & \underline{0.78}\\
Reflexion & 11.7 & 0.80 & 0.85 & 39.1 & 0.85 & 0.92 & 20.3 & 0.85 & 0.93 & 8.8 & 0.82 & 0.86 & 16.6 & 0.79 & 0.85 & 4.8 & 0.85 & 0.91\\
GEPA & 12.9 & \underline{0.70} & \underline{0.77} & \textbf{44.4} & \textbf{0.68} & \textbf{0.72} & \textbf{25.1} & \textbf{0.68} & \textbf{0.76} & 11.3 & \underline{0.65} & 0.75 & \textbf{20.7} & \underline{0.67} & \textbf{0.72} & 8.2 & 0.77 & 0.85\\
Trace2Skill & 13.1 & 0.72 & \underline{0.77} & 44.0 & 0.77 & 0.82 & 24.1 & \underline{0.69} & \textbf{0.76} & \textbf{13.4} & \textbf{0.64} & \textbf{0.69} & 19.8 & \textbf{0.65} & \underline{0.73} & \textbf{9.3} & \textbf{0.68} & \textbf{0.77}\\
\model{} & \textbf{14.5} & \underline{0.70} & \textbf{0.75} & \underline{44.2} & \underline{0.72} & \underline{0.76} & \underline{25.0} & 0.74 & \underline{0.79} & \underline{13.2} & 0.69 & 0.74 & \underline{20.6} & 0.71 & 0.76 & 8.6 & 0.73 & \underline{0.78}\\
\midrule
\multicolumn{19}{l}{\textbf{Qwen3.5-9B}}\\
BoN-5 & 17.6 & 1.06 & 1.22 & 62.4 & 1.18 & 1.30 & 40.3 & 1.13 & 1.25 & 12.4 & 1.13 & 1.28 & 28.1 & 1.16 & 1.23 & 7.8 & 1.12 & 1.23\\
RTS & 25.2 & 1.00 & 1.00 & 64.8 & 1.00 & 1.00 & 48.2 & 1.00 & 1.00 & \underline{19.5} & 1.00 & 1.00 & 32.6 & 1.00 & 1.00 & \underline{14.1} & 1.00 & 1.00\\
ACE & 23.8 & \underline{0.63} & \textbf{0.68} & 63.6 & 0.73 & 0.83 & 46.0 & 0.76 & 0.82 & \textbf{19.7} & \underline{0.62} & \underline{0.68} & 31.9 & \textbf{0.64} & \textbf{0.72} & 13.6 & \textbf{0.63} & \textbf{0.69}\\
ReasoningBank & \underline{25.4} & 0.69 & 0.73 & 64.5 & 0.71 & 0.75 & 46.5 & \textbf{0.62} & \textbf{0.68} & 16.7 & 0.69 & 0.77 & 31.7 & 0.72 & 0.81 & 13.2 & 0.73 & 0.77\\
Reflexion & 24.2 & 0.77 & 0.84 & 62.1 & 0.77 & 0.84 & 43.9 & 0.80 & 0.88 & 16.6 & 0.79 & 0.87 & 31.5 & 0.82 & 0.88 & 9.5 & 0.78 & 0.88\\
GEPA & 24.4 & 0.70 & 0.79 & \textbf{65.3} & 0.69 & 0.78 & \textbf{48.9} & \underline{0.63} & \underline{0.71} & 18.5 & \textbf{0.58} & \textbf{0.62} & 31.2 & 0.69 & \underline{0.78} & 13.2 & 0.73 & 0.82\\
Trace2Skill & 24.5 & \textbf{0.61} & \textbf{0.68} & 63.5 & \textbf{0.61} & \textbf{0.69} & 47.7 & 0.71 & 0.78 & 19.0 & 0.64 & \underline{0.68} & \underline{32.9} & 0.69 & \underline{0.78} & 13.3 & 0.70 & 0.79\\
\model{} & \textbf{25.6} & 0.65 & \underline{0.70} & \underline{65.0} & \underline{0.66} & \underline{0.72} & \underline{48.5} & 0.69 & 0.73 & 19.4 & 0.64 & 0.69 & \textbf{33.2} & \underline{0.67} & \textbf{0.72} & \textbf{14.3} & \underline{0.67} & \underline{0.71}\\
\midrule
\multicolumn{19}{l}{\textbf{Qwen3.5-27B}}\\
BoN-5 & 23.5 & 1.07 & 1.28 & 63.2 & 1.18 & 1.26 & 48.3 & 1.13 & 1.29 & 18.9 & 1.18 & 1.22 & 33.9 & 1.08 & 1.28 & 11.6 & 1.07 & 1.28\\
RTS & 28.7 & 1.00 & 1.00 & \underline{70.6} & 1.00 & 1.00 & 53.8 & 1.00 & 1.00 & \underline{22.1} & 1.00 & 1.00 & \underline{35.9} & 1.00 & 1.00 & \textbf{15.9} & 1.00 & 1.00\\
ACE & 27.8 & 0.64 & 0.73 & 68.5 & \underline{0.59} & 0.68 & 53.0 & 0.70 & \underline{0.74} & 20.4 & 0.68 & 0.78 & 33.7 & \underline{0.59} & \textbf{0.64} & 14.4 & \underline{0.66} & \underline{0.72}\\
ReasoningBank & 27.8 & 0.65 & 0.75 & 70.0 & 0.64 & 0.70 & 51.3 & \textbf{0.64} & \underline{0.74} & 21.2 & \textbf{0.59} & \underline{0.68} & 33.5 & 0.67 & 0.74 & 14.9 & \textbf{0.64} & 0.75\\
Reflexion & 26.2 & 0.78 & 0.89 & 68.8 & 0.72 & 0.82 & 49.2 & 0.78 & 0.88 & 19.5 & 0.75 & 0.82 & 33.7 & 0.70 & 0.77 & 13.1 & 0.78 & 0.83\\
GEPA & \underline{28.8} & \textbf{0.60} & \underline{0.70} & 69.4 & \textbf{0.57} & \textbf{0.64} & \textbf{54.3} & \underline{0.65} & \underline{0.74} & 21.2 & 0.66 & 0.72 & 35.3 & 0.60 & 0.68 & 14.5 & 0.70 & 0.81\\
Trace2Skill & \textbf{29.2} & 0.66 & 0.74 & 69.5 & 0.65 & 0.70 & 53.7 & 0.66 & 0.76 & 20.8 & 0.68 & 0.74 & 35.5 & \textbf{0.58} & \underline{0.66} & 15.0 & \underline{0.66} & \underline{0.72}\\
\model{} & \textbf{29.2} & \underline{0.61} & \textbf{0.66} & \textbf{71.0} & 0.63 & \underline{0.67} & \underline{54.0} & 0.66 & \textbf{0.70} & \textbf{22.5} & \underline{0.62} & \textbf{0.66} & \textbf{36.4} & 0.62 & \underline{0.66} & \underline{15.7} & \textbf{0.64} & \textbf{0.70}\\
\midrule
\multicolumn{19}{l}{\textbf{Qwen3.5-122B-A10B}}\\
BoN-5 & 25.1 & 1.08 & 1.20 & 66.0 & 1.14 & 1.25 & 49.1 & 1.11 & 1.23 & 19.5 & 1.16 & 1.24 & 34.9 & 1.09 & 1.22 & 12.1 & 1.13 & 1.25\\
RTS & \underline{29.8} & 1.00 & 1.00 & \underline{72.0} & 1.00 & 1.00 & 53.4 & 1.00 & 1.00 & \underline{23.2} & 1.00 & 1.00 & 37.0 & 1.00 & 1.00 & \underline{16.6} & 1.00 & 1.00\\
ACE & 28.6 & \underline{0.63} & 0.72 & 70.9 & \underline{0.61} & 0.69 & 52.4 & 0.69 & 0.76 & 21.6 & 0.66 & 0.75 & 35.6 & 0.62 & 0.69 & 15.4 & \underline{0.66} & \underline{0.73}\\
ReasoningBank & 28.8 & 0.65 & 0.74 & 71.5 & 0.62 & 0.71 & 51.9 & \underline{0.66} & 0.74 & 21.8 & \textbf{0.62} & \underline{0.71} & 35.2 & 0.65 & 0.73 & 15.1 & \textbf{0.64} & \underline{0.73}\\
Reflexion & 27.3 & 0.77 & 0.85 & 69.8 & 0.75 & 0.83 & 49.8 & 0.80 & 0.87 & 20.0 & 0.77 & 0.85 & 34.1 & 0.74 & 0.82 & 14.3 & 0.80 & 0.87\\
GEPA & 29.3 & \textbf{0.60} & \underline{0.69} & \textbf{72.4} & \textbf{0.59} & \textbf{0.66} & \textbf{53.6} & \textbf{0.65} & \underline{0.73} & 22.5 & \underline{0.64} & \underline{0.71} & 36.9 & \underline{0.61} & \underline{0.68} & 16.0 & 0.67 & 0.75\\
Trace2Skill & 29.5 & 0.64 & 0.72 & 71.6 & 0.62 & 0.70 & 53.1 & \underline{0.66} & 0.74 & 22.7 & 0.65 & 0.72 & \underline{37.1} & \textbf{0.60} & \textbf{0.67} & 16.3 & \textbf{0.64} & \textbf{0.71}\\
\model{} & \textbf{30.0} & \underline{0.63} & \textbf{0.68} & \textbf{72.4} & 0.62 & \underline{0.67} & \underline{53.5} & 0.68 & \textbf{0.72} & \textbf{23.4} & \textbf{0.62} & \textbf{0.67} & \textbf{37.3} & 0.64 & 0.69 & \textbf{16.8} & \underline{0.66} & \textbf{0.71}\\
\bottomrule
\end{tabular}
\endgroup
\end{table}

\subsection{Consistent Efficiency across Various Benchmarks}\label{sec:q1}

Across six diverse agentic benchmarks, \model{} consistently achieves a favorable quality-efficiency tradeoff, substantially reducing retrospective-search cost while preserving task quality within Table~\ref{tab:fp_main}. On average, \model{} reduces raw and uncached inference tokens by 36.2\% and 30.3\%, respectively, while slightly improving task score by $+0.23$ points relative to full RTS. The gains remain consistent across heterogeneous environments: benchmark-level raw-token savings range from 31.8\% on SWE-bench to 39.8\% on AutomationBench, while the corresponding mean score changes remain within $+0.50$ points. Overall, \model{} removes roughly one-third of retrospective inference across diverse agentic tasks without sacrificing overall task performance.





\subsection{Robust Efficiency across Different Backbone Scales}\label{sec:q2}

As illustrated in Table~\ref{tab:fp_main2}, across backbone scales, \model{} consistently preserves task quality while reducing retrospective-search cost. As the backbone scales from 4B to 9B and 27B, average raw-token savings increase from 28.5\% to 33.7\% and 37.0\%, while uncached-token savings increase from 23.7\% to 28.8\% and 32.5\%. At 122B-A10B, the savings remain comparable at 35.8\% and 31.0\%, indicating that the efficiency benefit largely saturates at the largest scale rather than disappearing. Meanwhile, mean score changes relative to full RTS remain stable at $+0.27$, $+0.27$, $+0.30$, and $+0.23$ points across the four scales. These results show that \model{}'s quality-efficiency advantage is not tied to a particular model size and persists from small to substantially larger backbones.

\subsection{Consistent Efficiency on Frontier Black-box Models}\label{sec:q3}

\begin{table}[h!]
\centering
\caption{
Black-box evaluation on proprietary backbones. $S$ is task score (0-100); $R/U$ are raw/uncached inference-token ratios to full RTS. Lower $R/U$ indicates lower deployment cost. The best is in \textbf{Bold} and the second best is in \underline{underline}.
}
\label{tab:blackbox}
\begingroup
\scriptsize
\setlength{\tabcolsep}{4.0pt}
\begin{tabular}{@{}l*{12}{r}@{}}
\toprule
& \multicolumn{6}{c}{GPT-5.6 Sol}
& \multicolumn{6}{c}{Opus 4.8} \\
\cmidrule(lr){2-7}\cmidrule(l){8-13}
& \multicolumn{3}{c}{Automation}
& \multicolumn{3}{c}{$\tau^3$-Bench}
& \multicolumn{3}{c}{Automation}
& \multicolumn{3}{c}{$\tau^3$-Bench} \\
\cmidrule(lr){2-4}\cmidrule(lr){5-7}
\cmidrule(lr){8-10}\cmidrule(l){11-13}
Method
& $S$ & $R$ & $U$
& $S$ & $R$ & $U$
& $S$ & $R$ & $U$
& $S$ & $R$ & $U$ \\
\midrule
BoN-5
& 59.83 & 2.030 & 1.689
& 59.28 & 0.989 & 1.138
& 55.17 & 1.491 & 1.025
& 44.44 & 1.137 & 2.743 \\
RDS
& 68.83 & 1.632 & 0.697
& 55.67 & 2.783 & 0.775
& 62.50 & 3.188 & 1.192
& 29.90 & 1.685 & 0.454 \\
GHCP
& 33.50 & 0.846 & 0.481
& 43.30 & 0.208 & 0.186
& 39.83 & 0.487 & 0.189
& 41.24 & 0.171 & 0.251 \\
RTS
& \textbf{71.83} & 1.000 & 1.000
& \textbf{67.53} & 1.000 & 1.000
& \textbf{70.83} & 1.000 & 1.000
& \underline{44.67} & 1.000 & 1.000 \\
\model{}
& \underline{69.00} & 0.764 & 0.556
& \underline{64.43} & 0.588 & 0.408
& \underline{66.67} & 0.778 & 0.763
& \textbf{48.31} & 0.611 & 0.569 \\
\bottomrule
\end{tabular}
\endgroup
\end{table}

Deployed in a production agentic system where the agent uses GPT-5.6 Sol and Claude Opus 4.8 served through production inference platforms employing state-of-the-art caching, \model{} retains substantial efficiency gains (Table~\ref{tab:blackbox}). Across AutomationBench and $\tau^3$-Bench, it uses only 58.8-77.8\% of RTS raw tokens and 40.8-76.3\% of uncached tokens, corresponding to 22.2-41.2\% and 23.7-59.2\% savings, respectively, while remaining comparable in performance to RTS. The gains are not reproduced by simply extending a single retrospective trajectory: compared with RDS, \model{} achieves higher task score in all four settings while reducing raw inference cost by 53.1-78.9\%. It also exceeds BoN-5 in task quality investing much less tokens. GHCP occupies a substantially cheaper but lower-quality operating point in three settings. Overall, the results show that \model{}'s adaptive retrospective allocation transfers to frontier black-box backbones, preserving much of full-search quality while substantially reducing token consumption. Appendix~\ref{app:addresult} reports the corresponding absolute inference-token totals and RTS-normalized costs for all black-box settings.
In our deployment, \model{} also demonstrates robustness to changes in rollout scheduling of host production agentic system while achieving further gains in token efficiency (Appendix~\ref{app:rollout_scheduling}).

\subsection{Leakage-Free Feedback and Sample-Efficient Self-Evolution}\label{sec:q4}

\textbf{Feedback-leakage audit.}
We examine whether the observed performance depends on evaluator feedback in the retrospective prompt for agent. The \emph{feedback-masked} only retains task execution history, but removes explicit rewards, rubric/grader-specific fields, and reference-answer cues. 
Across the four audited settings, 
feedback-masked scores are similar to the original (Table~\ref{tab:feedback_audit}). We also shuffle the feedback across tasks, introducing task-mismatched noise. Scores fall by 1.4-2.8 points relative to feedback-masked, yet remain broadly comparable across different backbone scales and variate benchmarks. These results indicate that the observed gains are not driven by evaluator-information leakage through retrospective prompt.

\begin{wraptable}{r}{0.48\linewidth}
\centering
\caption{
Feedback-leakage audit on two representative backbones. 
}
\label{tab:feedback_audit}
\begingroup
\scriptsize
\setlength{\tabcolsep}{4.5pt}
\begin{tabular}{@{}llrr@{}}
\toprule
Backbone & Feedback & Automation & $\tau^3$-Bench \\
\midrule
\multirow{3}{*}{Qwen3.5-9B}
& Original  & 14.5 & 44.2 \\
& Masked & 14.4 & 44.0 \\
& Shuffled  & 13.0 & 41.7 \\
\midrule
\multirow{3}{*}{Qwen3.5-27B}
& Original  & 25.6 & 65.0 \\
& Masked & 25.5 & 64.8 \\
& Shuffled  & 23.2 & 62.0 \\
\bottomrule
\end{tabular}
\endgroup
\end{wraptable}

\begin{figure}[h!]
\centering
\includegraphics[width=\linewidth]{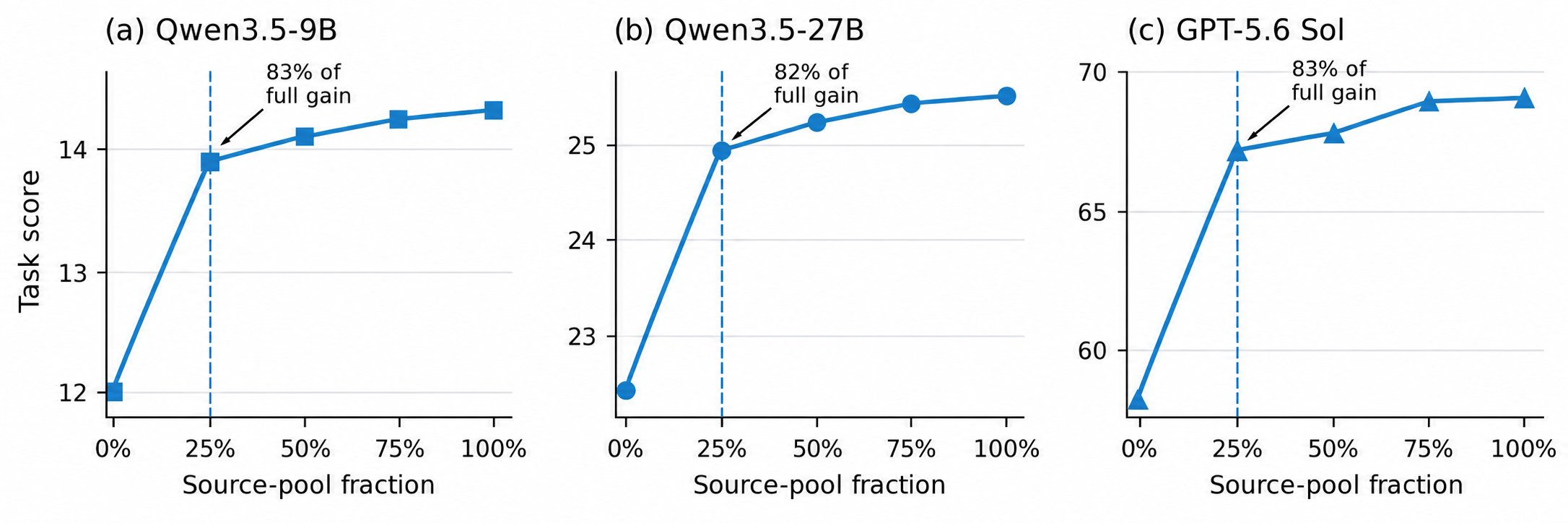}
\caption{Fraction of the full-training improvement recovered as the amount of training experience available for policy learning increases on Automation Bench. The percentages on the x-axis denote the fraction of the total training split used for policy learning.}
\label{fig:learning_curves}
\end{figure}

\textbf{Sample-efficient self-evolution.}
We also study the sample efficiency of \model{}'s self-evolution on Automation Bench by varying the fraction of the training split available for policy learning (Fig.~\ref{fig:learning_curves}). We report the fraction of the full-training improvement recovered, $R(f)=(S_f-S_0)/(S_{100}-S_0)$. Remarkably, with only 25\% of the training split(13.3\% of the total dataset), \model{} already captures 81-85\% of the improvement obtained using all training samples across the evaluated settings; the remaining 75\% contributes only another 15-19\%. This strongly front-loaded scaling shows that \model{} can rapidly bootstrap most of its eventual search-allocation improvement from limited completed experience, making its self-evolution highly sample efficient rather than dependent on a large training pool.

\section{Related Work}\label{sec:related_work}
\textbf{Retrospective Search and Adaptive Computation.}
Retrospective search improves reasoning by revisiting earlier decisions and exploring alternative trajectories. Tree of Thoughts~\citep{yao2023tree}, Language Agent Tree Search~\citep{zhou2024lats}, and ExACT~\citep{yu2025exact} instantiate this idea for reasoning and interactive agents. More broadly, metareasoning and resource-rational planning study which computations are worth performing under limited resources~\citep{russell1991metareasoning,hay2012selecting,callaway2018resource}. Recent LLM methods further adapt test-time computation according to query difficulty, intermediate confidence, or subproblem complexity, including compute-optimal scaling~\citep{snell2025scaling}, input-adaptive allocation~\citep{damani2025hard}, Certaindex~\citep{fu2025certaindex}, Learning When to Plan~\citep{paglieri2025learning}, and Plan-and-Budget~\citep{lin2026planbudget}. 

\textbf{Learning to Control Search.}
A complementary line of work learns from completed interactions to improve future reasoning. Reflexion~\citep{shinn2023reflexion}, ReasoningBank~\citep{ouyang2026reasoningbank}, ACE~\citep{zhang2026ace}, GEPA~\citep{agrawal2026gepa}, and Trace2Skill~\citep{ni2026trace2skill} use prior trajectories to construct reflections, memories, optimized context, or reusable skills. These methods primarily improve what information is available to the agent rather than learning how much retrospective computation should be allocated during execution. 
\section{Conclusion}\label{sec:conclusion}
Long-horizon, multi-call reasoning makes retrospective search a dynamic computation-allocation problem: delayed outcomes obscure the contribution of intermediate search decisions, while evolving execution evidence can make previously learned estimates stale. We characterize these challenges as \textit{Attribution Complexity} and \textit{Adaptation Complexity}, and introduce Search Value Dynamics (\method) to characterize the evolving trade-off between the gain and cost of retrospective search.
Building on \method, we propose \model{}, a self-evolving two-stage policy-learning framework. \model{} first learns a search-value policy offline from completed trajectories, then, in the second stage, makes search decisions on the current observations and then self-evolves, incorporating newly completed trajectories into subsequent policy updates. Across six benchmarks and multiple widely used LLM backbone families, \model{} achieves comparable state-of-the-art performance while reducing deployment-time token usage by up to 59.2\%.


\newpage
{\small
\bibliographystyle{plainnat}
\bibliography{iclr2027_conference}
}

\clearpage
\appendix

\section{Derivation and Scope of Search Value Dynamics}\label{app:proofs}

\subsection{Variance identities for paired search effects}
Fix a source input $z_{\mathrm{src}}$, a feasible action $u$, and the paired-replay rule $\Gamma$. As in Section~\ref{sec:svd_analysis}, let $D(u)=\Delta C(u)/\bar C_{\mathrm{src}}$, $\lambda_D=\lambda\bar C_{\mathrm{src}}$, and $W(u)=Y(u)-\lambda_D D(u)$. The first calculation conditions on the fixed source input and action under the source law $Q$. Whenever the component second moments exist, expansion of the variance gives
\begin{align}
V^\Gamma_{\mathrm{src}}(u)
&=\Var_Q[W(u)\mid z_{\mathrm{src}}]\notag\\
&=A^\Gamma_{\mathrm{src}}(u)
  +\lambda_D^2\Var_Q[D(u)\mid z_{\mathrm{src}}]
  -2\lambda_D\operatorname{Cov}_Q(Y(u),D(u)\mid z_{\mathrm{src}}),
\label{eq:app_net_variance}
\end{align}
where $A^\Gamma_{\mathrm{src}}(u)=\Var_Q[Y(u)\mid z_{\mathrm{src}}]$. Thus paired reward variation alone does not determine the sampling variation of the net effect. If $W_1(u),\ldots,W_n(u)$ are conditionally independent draws from this same law, their mean $\overline W_n(u)$ satisfies
\[
\E_Q[\overline W_n(u)\mid z_{\mathrm{src}}]
 =\E_Q[W(u)\mid z_{\mathrm{src}}],
\qquad
\Var_Q[\overline W_n(u)\mid z_{\mathrm{src}}]
 =\frac{V^\Gamma_{\mathrm{src}}(u)}{n}.
\]
Independence across \emph{actions} is not needed for this calculation or for the simultaneous bound below.

The other variance identity in Section~\ref{sec:svd_analysis} follows from the conditional law of total variance. Let $H$ be the permitted history and $Z=\phi(H)$, and fix an action feasible across the histories being compared. Writing $g(H,u)=\E_Q[W(u)\mid H]$, the fact that $Z$ is a function of $H$ implies
\begin{align}
\Var_Q[W(u)\mid Z]
&=\E_Q[\Var_Q(W(u)\mid H)\mid Z]
  +\Var_Q\!\bigl(g(H,u)\mid Z\bigr).
\label{eq:app_representation_variance}
\end{align}
The first term reflects variation remaining after the full permitted history is known; the second reflects different conditional values assigned the same controller input. Neither identity by itself implies that pairing reduces variance relative to a particular unpaired estimator.

\subsection{Proof of Theorem~\ref{thm:search_value_dynamics}}\label{app:svd_theorem_proof}
We use the notation and fixed-quantity conditions of Theorem~\ref{thm:search_value_dynamics}. In particular, $\mathcal F_t=\mathcal U_t\setminus\{u^0\}$ contains $K\geq 1$ actions, the source and current laws $Q,P$ use the same $\Gamma$, and the action set, inputs, and sample counts are fixed independently of the replay draws. For $L\in\{P,Q\}$, write $m_L(z,u)=\E_L[Y(u)\mid z]$ and $c_L(z,u)=\E_L[D(u)\mid z]$, as in the main text. Define the population source net effect
\[
g_Q(u)
=m_Q(z_{\mathrm{src}},u)-\lambda_D c_Q(z_{\mathrm{src}},u)
=\E_Q[W(u)\mid z_{\mathrm{src}}].
\]
All these conditional expectations are understood at inputs where their specified conditional laws are defined. The same feasible action and replay rule are used on both sides of each comparison.

\paragraph{Step 1: simultaneous source sampling error.}
For each $u\in\mathcal F_t$, let $\overline W_{\mathrm{src}}(u)$ denote the mean of its $n_u$ conditionally i.i.d.\ source paired net effects. Conditional on the fixed quantities in the theorem,
\[
\E[\overline W_{\mathrm{src}}(u)]=g_Q(u),
\qquad
\Var[\overline W_{\mathrm{src}}(u)]
  =\frac{V^\Gamma_{\mathrm{src}}(u)}{n_u}.
\]
When $V^\Gamma_{\mathrm{src}}(u)>0$, Chebyshev's inequality with $a_u=\sqrt{K V^\Gamma_{\mathrm{src}}(u)/(n_u\delta)}$ yields
\[
\Pr\!\left(
  \left|\overline W_{\mathrm{src}}(u)-g_Q(u)\right|>a_u
\right)
\leq \frac{V^\Gamma_{\mathrm{src}}(u)}{n_u a_u^2}
=\frac{\delta}{K}.
\]
If $V^\Gamma_{\mathrm{src}}(u)=0$, the sample mean equals $g_Q(u)$ almost surely, so the same event bound holds with $a_u=0$. A union bound over the $K$ actions therefore gives an event $\mathcal E_\delta$ with $\Pr(\mathcal E_\delta)\geq 1-\delta$ on which, simultaneously for all $u\in\mathcal F_t$,
\begin{equation}
\left|\overline W_{\mathrm{src}}(u)-g_Q(u)\right|
\leq \sqrt{\frac{K V^\Gamma_{\mathrm{src}}(u)}{n_u\delta}}.
\label{eq:app_simultaneous_sampling}
\end{equation}
The replay samples for different actions need not be independent. The source-fitted predictor may also depend on these samples: its realized discrepancy from their mean is retained explicitly as $e_{\mathrm{src}}(u) =|\widehat G_{\mathrm{src}}(u)-\overline W_{\mathrm{src}}(u)|$.

\paragraph{Step 2: change of input and law.}
Add and subtract the current-law reward mean at $z_{\mathrm{src}}$. Since $G_t(u)=m_P(z_t,u)-\lambda_Dc_P(z_t,u)$, the following is an exact identity:
\begin{align}
G_t(u)-g_Q(u)
&=\bigl[m_P(z_t,u)-m_P(z_{\mathrm{src}},u)\bigr]\notag\\
&\quad+\bigl[m_P(z_{\mathrm{src}},u)-m_Q(z_{\mathrm{src}},u)\bigr]\notag\\
&\quad-\lambda_D
  \bigl[c_P(z_t,u)-c_Q(z_{\mathrm{src}},u)\bigr].
\label{eq:appendix_decomp}
\end{align}
The first bracket has magnitude $\Omega_{\mathrm{src}\to t}(u)$; the third has magnitude $\xi_{\mathrm{src},t}(u)$. To bound the second bracket, let $P^u_{z_{\mathrm{src}}}$ and $Q^u_{z_{\mathrm{src}}}$ be the corresponding conditional laws of $Y(u)$. Because $R\in[0,1]$, the paired reward difference satisfies $Y(u)\in[-1,1]$. With $\mathrm{TV}(\mu,\nu)=\sup_A|\mu(A)-\nu(A)|$,
\begin{align}
|m_P(z_{\mathrm{src}},u)-m_Q(z_{\mathrm{src}},u)|
&=\left|\int y\,d\!\left(P^u_{z_{\mathrm{src}}}
                       -Q^u_{z_{\mathrm{src}}}\right)(y)\right|\notag\\
&\leq 2\,\mathrm{TV}
   \!\left(P^u_{z_{\mathrm{src}}},Q^u_{z_{\mathrm{src}}}\right)
 =S_{\mathrm{src}}(u).
\label{eq:app_tv_bound}
\end{align}
Consequently, by the triangle inequality and $\lambda_D\geq 0$,
\[
|G_t(u)-g_Q(u)|
\leq \Omega_{\mathrm{src}\to t}(u)
    +S_{\mathrm{src}}(u)
    +\lambda_D\xi_{\mathrm{src},t}(u).
\]
When $P^u_{z_{\mathrm{src}}}$ is absolutely continuous with respect to $Q^u_{z_{\mathrm{src}}}$ and the KL divergence is finite, Pinsker's inequality additionally gives $S_{\mathrm{src}}(u) \leq\sqrt{2D_{\mathrm{KL}} (P^u_{z_{\mathrm{src}}}\Vert Q^u_{z_{\mathrm{src}}})}$.

\paragraph{Step 3: combine the errors.}
On the single event $\mathcal E_\delta$, for every $u\in\mathcal F_t$,
\begin{align}
|G_t(u)-\widehat G_{\mathrm{src}}(u)|
&\leq |G_t(u)-g_Q(u)|
      +|g_Q(u)-\overline W_{\mathrm{src}}(u)|
      +|\overline W_{\mathrm{src}}(u)-\widehat G_{\mathrm{src}}(u)|\notag\\
&\leq e_{\mathrm{src}}(u)
      +\sqrt{\frac{K V^\Gamma_{\mathrm{src}}(u)}{n_u\delta}}
      +\Omega_{\mathrm{src}\to t}(u)\notag\\
&\quad+S_{\mathrm{src}}(u)
      +\lambda_D\xi_{\mathrm{src},t}(u)
 =\beta^\delta_{\mathrm{src},t}(u).
\label{eq:app_main_bound}
\end{align}
This proves Equation~\ref{thm:search_value_dynamics}. For any $\eta>0$, the condition
\[
n_u\geq \frac{K V^\Gamma_{\mathrm{src}}(u)}{\delta\eta^2}
\]
is sufficient to make the displayed sampling term for that action at most $\eta$; it does not control the other terms.

\subsection{Decision consequences and scope}
For non-skip actions, set $L_t^\delta(u)=\widehat G_{\mathrm{src}}(u) -\beta^\delta_{\mathrm{src},t}(u)$ and $U_t^\delta(u)=\widehat G_{\mathrm{src}}(u) +\beta^\delta_{\mathrm{src},t}(u)$, and extend both quantities to skip by $L_t^\delta(u^0)=U_t^\delta(u^0)=G_t(u^0)=0$. Equation~\ref{eq:app_main_bound} implies $L_t^\delta(u)\leq G_t(u)\leq U_t^\delta(u)$ simultaneously over $\mathcal U_t$ on $\mathcal E_\delta$.

If $u_L\in\arg\max_{u\in\mathcal F_t}L_t^\delta(u)$ and $L_t^\delta(u_L)>0$, then $G_t(u_L)\geq L_t^\delta(u_L)>0$; in particular, skip cannot maximize the true local net value. Let $u^\star\in\arg\max_{u\in\mathcal U_t}G_t(u)$, so $u^\star\in\mathcal F_t$. Because $L_t^\delta(u_L)\geq L_t^\delta(u^\star)$,
\begin{align}
G_t(u^\star)-G_t(u_L)
&\leq U_t^\delta(u^\star)-L_t^\delta(u_L)\notag\\
&\leq U_t^\delta(u^\star)-L_t^\delta(u^\star)
 =2\beta^\delta_{\mathrm{src},t}(u^\star).
\label{eq:app_local_regret}
\end{align}
If instead $\max_{u\in\mathcal F_t}U_t^\delta(u)\leq 0$, then $G_t(u)\leq 0=G_t(u^0)$ for every non-skip action. Skip is therefore a local maximizer, possibly tied with another action. These are precisely the positive-LCB, skip-optimality, and local-regret implications stated after Theorem~1.

For completeness, a related information inequality follows without a sampling argument. Let $\mathcal U$ be a fixed finite action set feasible for all histories under comparison, and let $g(H,u)$ be integrable for every $u\in\mathcal U$. Since $g(H,u)\leq\max_{v\in\mathcal U}g(H,v)$ for each fixed $u$, conditioning on $Z=\phi(H)$ and then maximizing gives
\[
\max_{u\in\mathcal U}\E[g(H,u)\mid Z]
\leq \E\!\left[\max_{u\in\mathcal U}g(H,u)\,\middle|\,Z\right].
\]
Taking expectations preserves the inequality. The comparison holds under the same history distribution and fixed menu; it only describes the possible value of more informative inputs.

Theorem~\ref{thm:search_value_dynamics} conditions on fixed inputs, menu, counts, and paired-replay rule. Its event need not cover menus or inputs chosen adaptively after inspecting these replay samples. Moreover, $e_{\mathrm{src}}(u)$ is a realized source-sample fitting discrepancy, not an independently validated bound on prediction error; it may be large. At deployment the controller evaluates its predictor at the refreshed input $z_t$, while Theorem~1 bounds reuse of $\widehat G_{\mathrm{src}}(u)$ at $z_{\mathrm{src}}$. Neither the theorem nor the information inequality certifies the learned operational margin or a closed-loop regret guarantee for the deployed policy, whose subsequent decisions may differ from the fixed continuation rule $\rho$.
\section{Model Configuration}\label{app:controller}
This section provides the action-level specification omitted from the system walkthrough. We fix a task and omit task index $i$. Source collection, feature construction, and controller fitting use the task boundaries and information permissions defined in Section~\ref{sec:preliminaries}.

\subsection{State and action representation}
Each permitted event is converted into categorical and scalar features: event type, tool identity, branch relations, success or error category, completed-branch score summaries, depth, search count, and accumulated token usage. Bounded deterministic features represent tool-argument and observation structure, including key names, value types, and length bins. Features are constructed from the prefix available at the recorded decision; later events never overwrite an earlier example.

A recurrent encoder maps the observable controller state to a representation $r_t=f_\psi(z_t)$. The host supplies the state, feasible-action mask, branch relations, and remaining resources. Action embeddings describe each host-provided search configuration. Resource features use source-only normalization. Feasibility is enforced from the host's actual reservation rule rather than the learned mean-cost prediction.

\begin{table}[H]
\centering\small
\caption{Default controller configuration. Host caps and selected trade-off parameters are recorded separately in each evaluation manifest.}
\label{tab:configuration}
\begin{tabularx}{\textwidth}{@{}lX@{}}
\toprule
Component & Setting\\
\midrule
Event embedding / recurrent state & 64 / 128 dimensions; one GRU layer\\
Prediction ensemble & 5 action-conditioned heads; task-bootstrap weights\\
Head architecture & One 64-unit hidden layer; separate gain and cost outputs\\
Source fitting & Adam, learning rate $10^{-3}$, batch size 64; inner-source validation stopping\\
Regularization & Gradient norm cap 1; task-balanced loss weights\\
Source partitions & 80\% predictor fit, 10\% margin fit, 10\% residual calibration, grouped by task\\
Action menu & $k\in\{1,k_{\max}\}$; $d\in\{\lceil d_{\max}/2\rceil,d_{\max}\}$; skip\\
Probe reservation & At most 10\% of remaining budget; one fork and short cap\\
Runtime support & $\ge20$ distinct source tasks and five-nearest-task distance at or below the source 90th percentile\\
Unsupported-state fallback & Source-selected Efficient RTS configuration under the same hard cap\\
Margin target quantile & $1-\delta_{\mathrm{cal}}=0.9$\\
Cost normalization & Source median host task cost, fixed at deployment\\
Trade-off selection & $\lambda\in\{0,0.01,0.03,0.1,0.3\}$ on source validation\\
\bottomrule
\end{tabularx}
\end{table}

\subsection{Paired replay protocol, targets, and value prediction}\label{app:paired_protocol}
Source states are sampled from a mixture of host budgets, including full-budget search, so that low apparent progress does not censor all recovery examples. The state rule is fixed before downstream reward is read. Candidate actions are sampled with logged nonzero probabilities. 

For each candidate action $u$, the sandbox restores the same state for $u$ and the skip reference $u^0$. Both arms use the same permissions, remaining budget, total-budget rule, and subsequent execution policy; either arm may search at later opportunities. Let $\bar C_{\mathrm{src}}>0$ be the cost scale fixed from source tasks.
For each paired replicate $q$, the targets are
\begin{equation}
\widetilde Y_q=R(\tau_q^{u})-R(\tau_q^{0}),
\qquad
\widetilde D_q=
\frac{C(\tau_q^{u})-C(\tau_q^{0})}{\bar C_{\mathrm{src}}}.
\label{eq:paired_targets}
\end{equation}
Each $C(\tau_q^a)$ includes computation used by all search branches generated within arm $a$, including branches that are not selected. When two continuations per arm are collected, corresponding replicate indices form two paired records.

A record stores the state identifier, permitted-prefix hash, candidate action and its sampling probability, paired outcomes and costs, replicate index, replay-protocol identifier, continuation-policy identifier, and source-task identifier. Protocol $\Gamma$ shares randomness only for exogenous events whose meaning and distribution do not depend on the chosen action. Such draws are keyed by task, state, replicate, and event identity. Backbone sampling and action-dependent environment draws use arm-specific keys, preserving the marginal execution law of each arm. Source collection and paired evaluation use the same replay protocol; controller fitting uses the resulting records. A pair is excluded from fitting if restoration fails to reproduce the state hash, permissions, or tool state, or if its effects cannot be isolated within the sandbox.

\begin{table}[H]
\centering\small
\caption{Paired-supervision collection statistics. Candidate action--skip pairs use two continuations per arm. Rejected pairs are excluded before fitting; restoration agreement compares restored decision state and permission hashes before either continuation runs.}
\label{tab:pair_collection}
\setlength{\tabcolsep}{4pt}
\begin{tabular*}{\textwidth}{@{\extracolsep{\fill}}lrrrrr@{}}
\toprule
Benchmark & Checkpoints & Candidates & Accepted & Reject (\%) & Restore (\%)\\\midrule
AutomationBench & 8,640 & 17,280 & 16,812 & 2.7 & 99.8\\
$\tau^3$-Bench & 6,480 & 12,960 & 12,532 & 3.3 & 99.7\\
SWE-bench Verified & 5,940 & 11,880 & 11,215 & 5.6 & 99.4\\
MM-ToolSandbox & 6,220 & 12,440 & 12,079 & 2.9 & 99.8\\
WorkArena++ & 5,760 & 11,520 & 10,967 & 4.8 & 99.5\\
VitaBench & 7,040 & 14,080 & 13,587 & 3.5 & 99.6\\
\bottomrule
\end{tabular*}
\end{table}

\paragraph{Isolated execution of parallel rollouts in real-world live environment.}
Parallel multi-turn rollouts can interfere through shared filesystem or database assets in a live real-world environment: writes by one rollout may alter another's observations, overwrite its changes, or invalidate subsequent tool operations. This cross-rollout interference also undermines the intended comparison between paired action and skip executions. Our production agentic system (host) therefore routes supported tool calls through a session-scoped virtualization layer. Each rollout receives a distinct session identifier and a private execution view; paired replay arms are initialized from the same base snapshot. Writes are staged in session-local overlays, and subsequent reads and tool outputs reflect the session's own updates without exposing other rollouts' writes. Retrospective forks inherit the parent's state at the chosen fork point while isolating subsequent modifications. Replay-generated changes remain sandboxed and are discarded after outcome collection. During production execution, only the final selected trajectory's staged changes are eligible for publication to live assets; unselected overlays are discarded. Publication requires validation of relevant state dependencies and permissions against the current live environment, with conflicts requiring revalidation or re-execution. Isolation covers only resources mediated by the virtualization layer, and commit guarantees follow the supported storage transaction boundaries. Irreversible external effects must be deferred or excluded from speculative execution.

For bootstrap head $j$, the fitting loss is
\begin{equation}
\mathcal L_j=\frac1{|\mathcal D_j|}
\sum_{q\in\mathcal D_j}
\Big[
\big(\widehat m^{(j)}(r_q,u_q)-\widetilde Y_q\big)^2
+
\big(\widehat c^{(j)}(r_q,u_q)-\widetilde D_q\big)^2
\Big].
\label{eq:training_loss}
\end{equation}
Whole tasks, rather than prefixes, are bootstrapped. The ensemble's
predicted net value is
\begin{equation}
\widehat G_t(u)=\frac1M\sum_{j=1}^{M}
\left[
\widehat m^{(j)}(r_t,u)-\lambda_D
\widehat c^{(j)}(r_t,u)
\right],
\qquad
\lambda_D=\lambda\bar C_{\mathrm{src}}.
\label{eq:controller_value}
\end{equation}
Here $\lambda$ is the raw-cost coefficient in Section~\ref{sec:preliminaries}, and $\widehat c^{(j)}$ predicts $\Delta C/\bar C_{\mathrm{src}}$. Head disagreement $s_t(u)$ is an empirical uncertainty feature, not a posterior standard deviation. Source-data collection cost is reported separately and counts both executed arms, failed runs, and state-preparation calls; these extra collection runs are not added to either arm's deployment cost $C(\tau_q^a)$.
Head disagreement $s_t(u)$ is an empirical uncertainty feature, not a posterior standard deviation. Ordinary unpaired traces can supply auxiliary targets, but they do not replace Eq.~\ref{eq:paired_targets} without an explicit attribution-bias term. Source collection cost includes every executed arm, failed run, and state-preparation call.

\subsection{Empirical margin and interval-consistent stopping rule}
Reference support is computed in source-standardized representation space for each action configuration. The novelty score $d_t(u)$ is the mean distance to nearby source representations from distinct tasks, and $n_t(u)$ counts distinct supporting tasks. A positive monotone calibrator forms
\begin{equation}
a_t(u)=\operatorname{softplus}\!\left(b+w_s s_t(u)+w_d d_t(u)+w_n[1+n_t(u)]^{-1/2}\right),
\quad w_s,w_d,w_n\ge0.
\label{eq:margin_features}
\end{equation}
It minimizes pinball loss at quantile $1-\delta_{\mathrm{cal}}$ on the margin-fitting split for the absolute paired net-target residual. On the residual-calibration split, one state-action pair is selected per task by a prespecified random rule. With $\widetilde W_q=\widetilde Y_q-\lambda\widetilde D_q/\bar C_{\mathrm{src}}$, define
\begin{equation}
r_q=|\widetilde W_q-\widehat G_q|-a_q,
\qquad
q_{\mathrm{cal}}=\max\!\left\{0,r_{(\lceil(N+1)(1-\delta_{\mathrm{cal}})\rceil)}\right\}.
\label{eq:calibration_quantile}
\end{equation}
The order statistic includes the conventional additional $+\infty$ value if its index exceeds $N$. Insufficient support therefore produces fallback instead of an artificially narrow margin. The resulting operational scores are
\begin{equation}
\widehat B_t(u)=a_t(u)+q_{\mathrm{cal}},\qquad
L_t(u)=\widehat G_t(u)-\widehat B_t(u),\qquad
U_t(u)=\widehat G_t(u)+\widehat B_t(u).
\label{eq:empirical_interval}
\end{equation}
Under exchangeability, this construction gives marginal coverage for a new \emph{realized paired target}. Because the residual also contains continuation randomness, this target differs from the conditional mean $G_t(u)$. Closed-loop action selection further changes the encountered state distribution; adaptive conformal inference studies related online distribution shift~\citep{gibbs2021adaptive}. Accordingly, $L_t$ and $U_t$ are operational decision scores whose closed-loop effect is measured through held-out rollouts and the no-margin ablation in Section~\ref{sec:q3}.

Let $\widehat m_t(u)$ and $\widehat c_t(u)$ denote the ensemble mean predictions. The lower and upper operational scores imply two distinct action-level boundaries:
\begin{equation}
\begin{aligned}
L_t(u)>0
&\Longleftrightarrow
\widehat m_t(u)>\lambda\widehat c_t(u)+\widehat B_t(u)
&&\text{(commit-eligible)},\\
U_t(u)>0
&\Longleftrightarrow
\widehat m_t(u)>\lambda\widehat c_t(u)-\widehat B_t(u)
&&\text{(probe-eligible)}.
\end{aligned}
\label{eq:action_boundaries}
\end{equation}
Both boundaries change with the current trajectory, candidate action, predicted cost, and empirical reliability; the global $\lambda$ remains source-selected. They are not interchangeable: the conservative $+\widehat B_t$ boundary authorizes regular execution, whereas the permissive $-\widehat B_t$ boundary only keeps a low-cost probe eligible. Equations~\ref{thm:search_value_dynamics}-\ref{eq:runtime_rule} aggregate these action-level tests. The system skips exactly when every feasible action fails the commit boundary and every feasible probe fails the probe boundary.

\subsection{Runtime rule and pseudocode}
The host provides low and full fork/depth configurations, removes duplicates, and masks infeasible actions through hard reservations. Equations~\ref{thm:search_value_dynamics}-\ref{eq:runtime_rule} select the highest positive-lower-score action, otherwise the least-cost positive-upper-score probe, and otherwise skip the current retrospective round. A skip does not end the task. If the artifact or support check is unavailable, the declared host fallback applies within the same hard budget.

\begin{algorithm}[h!]
\caption{\model{}: Closed-Loop Retrospective Search Allocation}
\label{alg:forkpilot}
\small
\begin{algorithmic}[1]
\Statex \textbf{Input:} retrospective-search host $\mathcal{H}$,
controller $\theta_n$, source-support index $\mathcal{S}$,
hard task budget $B$, trade-off coefficient $\lambda_D$
\State Initialize observable history $h_0$, search record $T_0$,
and remaining budget $b_0 \gets B$
\State $t \gets 0$
\While{the host has not terminated}
    \State Advance $\mathcal{H}$ to the next retrospective-search opportunity
    \State Append newly permitted events to $h_t$
    \State Obtain the host-feasible action menu $\mathcal{U}_t$
    \State $\mathcal{F}_t \gets \mathcal{U}_t \setminus \{u^0\}$
    \State Let $\mathcal{P}_t \subseteq \mathcal{F}_t$ denote bounded probe actions
    \State Form the observable state
    $z_t \gets \phi(h_t,T_t,b_t)$
    \State Compute representation $r_t \gets f_\psi(z_t)$
    \If{$\theta_n$ is unavailable \textbf{or}
        $\mathcal{S}(r_t,\mathcal{F}_t)$ fails}
        \State $u_t \gets
        \textsc{HostFallback}(\mathcal{H},\mathcal{U}_t,b_t)$
    \Else
        \For{$u \in \mathcal{F}_t$}
            \State Predict reward effects
            $\{\widehat m_t^{(j)}(u)\}_{j=1}^{M}$
            \State Predict normalized cost effects
            $\{\widehat c_t^{(j)}(u)\}_{j=1}^{M}$
            \State
            $\displaystyle
            \widehat G_t(u)
            \gets
            \frac{1}{M}
            \sum_{j=1}^{M}
            \left[
            \widehat m_t^{(j)}(u)
            -
            \lambda_D\widehat c_t^{(j)}(u)
            \right]$
            \State Compute empirical margin $\widehat B_t(u)$
            \State
            $L_t(u) \gets
            \widehat G_t(u)-\widehat B_t(u)$
            \State
            $U_t(u) \gets
            \widehat G_t(u)+\widehat B_t(u)$
        \EndFor
        \State
        $\displaystyle
        C_t \gets \max_{u\in\mathcal{F}_t} L_t(u)$
        \State
        $\displaystyle
        Q_t \gets \max_{u\in\mathcal{P}_t} U_t(u)$
        \If{$C_t > 0$}
            \State
            $\displaystyle
            u_t \gets
            \arg\max_{u\in\mathcal{F}_t} L_t(u)$
        \ElsIf{$Q_t > 0$}
            \State
            $\displaystyle
            u_t \gets
            \arg\min_{
            \substack{
            u\in\mathcal{P}_t\\
            U_t(u)>0}}
            \widehat c_t(u)$
        \Else
            \State $u_t \gets u^0$
            \Comment{skip this retrospective round}
        \EndIf
    \EndIf
    \State
    $(h_{t+1},T_{t+1},b_{t+1})
    \gets
    \textsc{Execute}(\mathcal{H},u_t)$
    \State $t \gets t+1$
\EndWhile
\If{the completed task is eligible source experience}
    \State Add its valid paired records to
    $\mathcal{D}_{n+1}$
    \State
    $\theta_{n+1}
    \gets
    \textsc{Learn}(\mathcal{D}_{n+1})$
\EndIf
\State Freeze the fitted controller before evaluating the next held-out block
\end{algorithmic}
\end{algorithm}

\begin{table}[H]
\centering\scriptsize
\caption{One end-to-end decision record (AutomationBench held-out trace AB-f2-0174). $\widehat G$, $\widehat B$, and $L$ refer to the selected non-skip candidate; the probe at $t=1$ is chosen by its positive upper score. The trace illustrates that a skip at $t=3$ does not prevent renewed search after later evidence.}
\label{tab:decision_trace}
\setlength{\tabcolsep}{2.5pt}
\begin{tabularx}{\textwidth}{@{}rXlrrrr@{}}
\toprule
$t$ & Newly visible evidence & Feasible actions & $\widehat G$ & $\widehat B$ & $L$ & Decision\\\midrule
1 & Lookup failed; no canonical record ID; 68\% budget remains & probe / full / skip & 0.018 & 0.026 & -0.008 & Probe\\
2 & Probe returns canonical ID and matching account & full / low / skip & 0.091 & 0.024 & 0.067 & Full\\
3 & Recovered branch raises permitted score from .61 to .83 & full / low / skip & 0.007 & 0.019 & -0.012 & Skip\\
4 & Later tool return exposes a recoverable permission path & full / low / skip & 0.054 & 0.021 & 0.033 & Full\\
\bottomrule
\end{tabularx}
\end{table}

\subsection{Online updates and censoring}
A task can enter the next fitting block only after all required downstream outcomes are complete. Online replay calls are logged as learning cost. A retained full-budget source panel preserves recovery opportunities that an adaptive policy might otherwise stop observing. No-future-improvement labels from truncated executions are censored rather than treated as zero search value.

Task roles for predictor fitting, margin fitting, and residual calibration remain disjoint and stable as the source prefix grows. The encoder, heads, normalization, support index, and margin are versioned and frozen as one artifact before held-out evaluation. Within a task, changing tool observations refresh the controller input and decision scores, but do not update these fitted parameters. Across completed source tasks, a later artifact may be fitted and deployed to a subsequent block. Inner source validation can select update hyperparameters; the outer test panel cannot determine whether an update is deployed. At cold start, none of these fitted artifacts or statistics is available, so the system uses the host fallback until fitting and calibration become possible.

\subsection{Baseline definitions}\label{app:baseline_details}
\paragraph{RTS and RDS.}
Full Retrospective Tree Search (RTS) uses the common retrospective-search host with a configuration selected on source validation to favor task score. Its three beams retain alternative continuations from earlier resumable states for exploration. The common scorer prioritizes promising branches for further expansion, providing exploitation within the search. At each eligible state, RTS executes the complete selected rollout allocation, subject to the hard host cap shared by all search-control methods.
Retrospective Depth Search (RDS) reviews one initial rollout and continues only from its final resumable node. It does not fork from earlier states. RDS therefore tests how much additional depth along a single trajectory can achieve without exploring alternative retrospective branches.

The eight comparators in the seven-backbone matrix are BoN-5, RTS, RDS, ACE, ReasoningBank, Reflexion, GEPA, and Trace2Skill. BoN-5 samples five complete candidates and selects by the common scorer. ACE, ReasoningBank, Reflexion, GEPA, and Trace2Skill use their declared context, memory, or prompt intervention while sharing the evaluation harness and token accounting. These are end-to-end baselines with a different intervention interface from rollout allocation; Table~\ref{tab:composition} evaluates their complementarity through a ReasoningBank+$\,$\model{} composition. BoN-1 and GHCP appear only in the separate frontier-model comparison and are not counted among the eight matrix comparators.

\begin{table}[H]
\centering\scriptsize
\caption{Composition with ReasoningBank on AutomationBench. The combined system supplies ReasoningBank context to the actor while \model{} independently controls retrospective fork/depth allocation. Ratios are relative to RTS; paired intervals compare the composition against \model{} alone.}
\label{tab:composition}
\setlength{\tabcolsep}{2.2pt}
\begin{tabularx}{\textwidth}{@{}l>{\raggedright\arraybackslash}Xrrrcc@{}}
\toprule
Backbone & Method & Score & $R$ & $U$ & $\Delta S$ [95\% CI] & $\Delta R$ [95\% CI]\\
\midrule
Qwen3.5-9B & ReasoningBank & 12.50 & 0.690 & 0.780 & -- & --\\
 & \model{} & 14.53 & 0.700 & 0.750 & -- & --\\
 & ReasoningBank + \model{} & 14.91 & 0.586 & 0.663 & +0.38 [+0.09,+0.71] & -0.114 [-0.132,-0.098]\\
\midrule
Qwen3.5-27B & ReasoningBank & 25.40 & 0.690 & 0.730 & -- & --\\
 & \model{} & 25.61 & 0.650 & 0.700 & -- & --\\
 & ReasoningBank + \model{} & 26.08 & 0.552 & 0.621 & +0.47 [+0.12,+0.83] & -0.098 [-0.117,-0.080]\\
\bottomrule
\end{tabularx}
\end{table}

\section{Evaluation Protocol and Accounting}\label{app:protocol}
\subsection{Released model checkpoints}\label{app:models}
Table~\ref{tab:open_models} maps every backbone label to a public weight release. No proprietary backbone is part of the evaluation matrix. The controller uses observable events and does not inspect model activations or gradients. Each run must pin the release revision, tokenizer, serving configuration, reasoning setting, decoding parameters, and tool-call serialization; an unpinned model family name is insufficient for reproduction.

\begin{table}[h!]
\centering\small
\caption{Open-weight backbone identities. Licenses refer to the listed releases. DeepSeek-R1-671B denotes the full original R1 release, not a distilled variant.}
\label{tab:open_models}
\setlength{\tabcolsep}{3pt}
\begin{tabularx}{\textwidth}{@{}lXl@{}}
\toprule
Backbone & Public checkpoint & Release license\\
\midrule
DeepSeek-V4-Flash & \texttt{deepseek-ai/DeepSeek-V4-Flash}~\citep{fp_v4flash} & MIT\\
Kimi-K3 & \texttt{moonshotai/Kimi-K3}~\citep{fp_kimik3} & Kimi K3 License\\
DeepSeek-R1-671B & \texttt{deepseek-ai/DeepSeek-R1}~\citep{fp_r1} & MIT\\
Qwen3.5-4B & \texttt{Qwen/Qwen3.5-4B}~\citep{fp_qwen354} & Apache 2.0\\
Qwen3.5-9B & \texttt{Qwen/Qwen3.5-9B}~\citep{fp_qwen359} & Apache 2.0\\
Qwen3.5-27B & \texttt{Qwen/Qwen3.5-27B}~\citep{fp_qwen3527} & Apache 2.0\\
Qwen3.5-122B-A10B& \texttt{Qwen/Qwen3.5-122B-A10B}~\citep{fp_qwen35122} & Apache 2.0\\
Qwen3.8-27B & \texttt{Qwen/Qwen3.8-27B}~\citep{fp_qwen3827} & Apache 2.0\\
\bottomrule
\end{tabularx}
\end{table}

For image-bearing observations, the shared adapter is a frozen Qwen3.5-9B instance with a fixed prompt that serializes only the public observation into text and structured tool fields. Every evaluated backbone, including those with native visual capability, receives this same adapter interface. The adapter cannot read answers, grader state, or a future trajectory; it is not trained on held-out tasks. Its input and output tokens count toward each method's cost, including unsuccessful branches. The manifest records the adapter checkpoint and prompt hash. This design permits a consistent all-open-weight comparison including text-only backbones, but its visual-task scores describe the adapted interface and must not be conflated with native-interface benchmark results.

\subsection{Task groups, outer folds, and source roles}
Every comparison is tied to a manifest containing benchmark revision, exact task identifiers, task groups, environment seeds, model checkpoint revisions, adapter configuration, host-search settings, cache policy, feedback permissions, and controller artifact identifier. The label \fptaubench{} identifies the text configuration used in the matrix; its suite revision and evaluator settings must be fixed in the manifest rather than inferred from the label alone. AutomationBench's public evaluation panel is distinct from any private leaderboard.

\begin{table}[H]
\centering\small
\caption{Evaluation manifest summary. Source/test counts are per outer fold; $k_{\max}$ and $d_{\max}$ are host fork and continuation-depth caps; the hard budget is relative to full RTS. All runs use the source-selected Efficient RTS fallback and the common support rule stated in text.}
\label{tab:manifest}
\begin{tabular*}{\textwidth}{@{\extracolsep{\fill}}lrrrrr@{}}
\toprule
Benchmark & Source & Test & $k_{\max}$ & $d_{\max}$ & Hard cap\\\midrule
AutomationBench & 400 & 200 & 4 & 6 & 1.00\\
$\tau^3$-Bench & 320 & 160 & 4 & 5 & 1.00\\
SWE-bench Verified & 300 & 200 & 3 & 6 & 1.00\\
MM-ToolSandbox & 320 & 160 & 4 & 5 & 1.00\\
WorkArena++ & 300 & 150 & 3 & 6 & 1.00\\
VitaBench & 400 & 200 & 4 & 5 & 1.00\\
\bottomrule
\end{tabular*}
\end{table}

The common host offers a retrospective decision after a completed search branch, a recoverable tool error, or an explicit host retrospection request. It applies the $k_{\max}$ and $d_{\max}$ settings in Table~\ref{tab:manifest} before presenting feasible actions. The controller support check requires at least 20 distinct source tasks for that action and a five-nearest-task representation distance no greater than the source 90th percentile. Failure of either condition invokes the recorded Efficient RTS fallback. These trigger, cap, support, and fallback settings are frozen from source validation.

For AutomationBench, the 600-task panel is divided into three grouped outer folds. Each run uses 400 tasks as its source pool and the remaining 200 as its held-out panel. Near-duplicate instances, common cloned states, all prefixes, and both replay arms remain within a task group. The same outer split is shared by every method and source-size condition. The aggregate contains out-of-fold predictions, not predictions on tasks used to train the corresponding controller. An instance can be a source task in another independent fold, but artifacts, memories, and thresholds cannot be transferred across those folds. No score from one fold selects another fold's controller.

Within each source pool, task roles are 80\% predictor fitting, 10\% margin fitting, and 10\% residual calibration. Hyperparameter and manual-threshold selection use inner task-grouped validation inside the predictor-fitting portion; they do not consume the residual-calibration or outer test panel. Role assignments are made before outcomes are read and preserved as nested source prefixes grow. All roles count toward $N$. At $N=100$, the allocation is 80/10/10 tasks, not 100 fitting tasks plus unreported calibration experience. Before fitting is possible, the fallback needs no source statistics.

The other benchmarks use the task-grouped outer counts in Table~\ref{tab:manifest}. Training never uses the scored task's group. Such within-benchmark learning is distinct from frozen cross-environment or unseen-backbone transfer; neither transfer claim follows from the main matrix. For any separate transfer analysis, encoder, heads, normalization, action semantics, support representation, margin, and $\lambda$ must be frozen together and fallback rates reported.

\subsection{Token costs, learning cost, and amortization}
\label{app:token_costs}
For task $i$ and complete-method repetition $r$, let $I_{ir}$, $O_{ir}$, and $H_{ir}$ be input, output, and cache-hit input tokens, summed across actor, search, permitted grading, observation adaptation, and auxiliary calls. Then
$
C^{\mathrm{raw}}_{ir}=I_{ir}+O_{ir},\qquad
C^{\mathrm{uncached}}_{ir}=I_{ir}-H_{ir}+O_{ir}.
$
Unknown cache-hit counts are missing data, not zeros. For method $m$ and reference $b$, the ratios are
$
R_m=\frac{\sum_{i,r}C^{\mathrm{raw},m}_{ir}}{\sum_{i,r}C^{\mathrm{raw},b}_{ir}},\qquad
U_m=\frac{\sum_{i,r}C^{\mathrm{uncached},m}_{ir}}{\sum_{i,r}C^{\mathrm{uncached},b}_{ir}}.
$
They are ratios of totals, not averages of task ratios. The Q1 RTS denominator remains fixed across source-size checkpoints and early-stopping settings. Cache policy is shared across methods, and every branch is charged regardless of whether it is ultimately selected.

Let $C_{\mathrm{src}}$ include source collection and all required paired continuations, and let $C_{\mathrm{update}}$ include additional online replay calls. For $N_{\mathrm{deploy}}$ deployed tasks, learning-inclusive average raw inference cost is
\begin{equation}
\overline C=\frac{C_{\mathrm{src}}+C_{\mathrm{update}}+
\sum_{i=1}^{N_{\mathrm{deploy}}}C_i^{\mathrm{deploy}}}{N_{\mathrm{deploy}}}.
\end{equation}
With positive per-task deployment savings $\Delta\overline C$ and fixed learning costs, the approximate token break-even point is $(C_{\mathrm{src}}+C_{\mathrm{update}})/\Delta\overline C$. Update frequency changes the numerator. 


\subsection{Robustness to Host Rollout Scheduling}
\label{app:rollout_scheduling}

The host controls when trajectories are launched, while \model{} allocates retrospective search during their execution. We examine whether its efficiency benefits persist when the host changes from three parallel trajectories to sequential execution, evaluating both RTS and \model{} under each schedule. Parallel execution explores diverse action sequences early, potentially finding a trajectory that requires fewer corrections. Sequential execution instead launches another trajectory only after abandoning the current attempt without solving the task, avoiding unnecessary launches when one trajectory suffices. Both schedules retain retrospective search within each trajectory, but sequential execution may increase latency when several attempts are needed.

\begin{table}[h!]
\centering
\caption{Host-scheduling comparison with GPT-5.6 Sol at extra-high reasoning effort. $S$ is task score (0--100); all $R/U$ values are raw/uncached deployment-token ratios to RTS (parallel) within each benchmark. Higher $S$ and lower $R/U$ are better. Bold and underline indicate the best and second-best values per column.}
\label{tab:rollout_scheduling}
\begingroup
\scriptsize
\setlength{\tabcolsep}{6pt}
\begin{tabular}{@{}l*{6}{r}@{}}
\toprule
Method & \multicolumn{3}{c}{AutomationBench}
       & \multicolumn{3}{c}{\fptaubench{}} \\
\cmidrule(lr){2-4}\cmidrule(lr){5-7}
& $S$ & $R$ & $U$ & $S$ & $R$ & $U$ \\
\midrule
RTS (parallel)
& 71.83 & 1.000 & 1.000
& \underline{67.53} & 1.000 & 1.000 \\
\model{} (parallel)
& 69.00 & 0.764 & \underline{0.556}
& 64.43 & \underline{0.588} & \underline{0.408} \\
\midrule
RTS (sequential)
& \textbf{77.00} & \underline{0.680} & 0.634
& \textbf{69.59} & 0.775 & 0.814 \\
\model{} (sequential)
& \underline{75.00} & \textbf{0.679} & \textbf{0.402}
& 64.95 & \textbf{0.505} & \textbf{0.292} \\
\bottomrule
\end{tabular}
\endgroup
\end{table}

\textbf{Sequential scheduling provides additional token savings.}
Both RTS and \model{} achieve higher observed scores with fewer tokens under sequential execution (Table~\ref{tab:rollout_scheduling}). Relative to parallel \model{}, its sequential variant reduces raw tokens by 11.1\% and 14.2\%, and uncached tokens by 27.6\% and 28.5\%, on AutomationBench and \fptaubench{}, respectively. Scores increase from 69.00 to 75.00 and from 64.43 to 64.95. Thus, deferring trajectory launches offers further savings beyond the default parallel configuration. Latency is not measured here; the potential delay from successive attempts remains unquantified.

\textbf{\model{} retains uncached-token savings under both host schedules.}
Against schedule-matched RTS, \model{} reduces uncached tokens by 44.4\%/59.2\% under parallel execution and 36.5\%/64.1\% under sequential execution on AutomationBench/\fptaubench{}. Under the sequential schedule, raw-token usage is nearly identical to RTS on AutomationBench and 34.8\% lower on \fptaubench{}, with respective score differences of $-2.00$ and $-4.64$ points. These operating points show that \model{}'s uncached-token efficiency benefit is not restricted to three-way parallel rollout, although the quality--cost trade-off varies with the host schedule.

\subsection{Scope of the frontier-model comparison}\label{app:frontier_scope}
The main frontier-model Pareto comparison uses GPT-5.6 Sol on AutomationBench and \fptaubench{} and considers only methods with both task score and uncached inference-token totals. A reported GEPA score without comparable token accounting cannot be assigned a Pareto rank. The separate completed Opus 4.8 (MAX) AutomationBench runs are reported in Table~\ref{tab:opus_automation} as a descriptive operating-point check. The raw and uncached ratios use full Opus RTS inference tokens as their respective denominators; source-training tokens are excluded from these deployment ratios.

\begin{table}[h!]
\centering\small
\caption{Completed Opus 4.8 (MAX) AutomationBench operating points. Scores are on the $[0,1]$ scale. $R/U$ are total raw/uncached inference-token ratios to full Opus RTS; lower token ratios are better. These aggregate records do not provide task-paired intervals.}
\label{tab:opus_automation}
\setlength{\tabcolsep}{8pt}
\begin{tabular}{@{}lrrr@{}}
\toprule
Method & Score $\uparrow$ & $R/\mathrm{RTS}\downarrow$ & $U/\mathrm{RTS}\downarrow$\\
\midrule
BoN-1 & 0.4233 & 0.295 & 0.195\\
BoN-5 & 0.5517 & 1.491 & 1.025\\
RTS & 0.7083 & 1.000 & 1.000\\
Efficient RTS & 0.6067 & 0.513 & 0.547\\
\model{} & 0.6667 & 0.778 & 0.763\\
\bottomrule
\end{tabular}
\end{table}

At this operating point, \model{} uses 22.2\% fewer raw and 23.7\% fewer uncached inference tokens than Opus RTS, with a 0.0416 lower task score. Its reported source-training totals are 1.541 billion raw and 0.457 billion uncached tokens and are not amortized in Table~\ref{tab:opus_automation}. No complete second-benchmark Opus comparison is available in the supplied run log, so no additional Opus Pareto claim is made. The aggregate Opus records also do not establish a noninferiority interval or a matched-cost superiority result.

\section{Self-evolving and Stopping Protocols}\label{app:learning}
\subsection{True cold start and nested source-size curves}
Each of five Q2 source orders is drawn independently of task outcomes and shared across training regimes; three controller seeds are fit per order. Nested prefixes contain 0, 100, 200, 300, and 400 completed source tasks per outer fold. At zero, there are no trained weights, calibration residuals, source normalizers, or experience-memory entries. Every request is handled by the host fallback under the predeclared deployment cap. A hand-initialized predictor with previously calibrated thresholds would instead constitute a warm start and is excluded from the cold-start condition.

The Q2 deployment cap is held fixed across all source sizes and regimes within a backbone and uses the same host reservation rule as Q1. The full-source batch endpoint is therefore directly aligned with the corresponding Q1 \model{} operating point rather than being evaluated under a second, undocumented cap. Training and replay budgets are logged separately from deployment cost.

For \emph{batch fitting}, the predictor is initialized afresh at each nonzero $N$ and fitted on that prefix's fitting tasks, with its margin and residual correction trained only on their assigned roles. For \emph{online updates}, the first nonzero artifact is fitted after the first completed block; subsequent artifacts warm-start from the preceding version and update on the expanded prefix, including retained fitting tasks. Recalibration uses all eligible completed tasks in the fixed margin and calibration roles. The source self-evolving policy used for paired targets remains the same frozen host across regimes, so different target definitions do not explain a curve difference.

Any paired replay evidence exposed to one learned regime is also available to the other at the corresponding source-size state. The comparison therefore isolates incremental fitting under an expanding source pool with matched evidence availability. Each artifact is frozen before the common held-out panel is evaluated. Held-out tasks and their feedback never enter updates, and the state schedule does not depend on their scores. The frozen cold-start control retains its initial fallback without updates.




\subsection{Saturation and order sensitivity}
Because the mean curves are mildly nonmonotone, define
\begin{equation}
N_{0.9}=\min\left\{N>0:\frac{S(N)-S(0)}{\max_{N'\in\mathcal N}S(N')-S(0)}\ge0.9\right\}.
\label{eq:saturation}
\end{equation}
Because the displayed mean curves can reverse slightly at larger $N$, Eq.~\ref{eq:saturation} normalizes by the best observed state on the evaluated source-size grid rather than by $S(N_{\max})$. On the mean curves, the first 90\% crossings are $N=300$ for 9B batch, $N=300$ for 9B online, $N=200$ for 27B batch, and $N=300$ for 27B online. We therefore use the curves to support early, noisy saturation rather than a universal 200-task threshold. Source-order and controller-seed variation remains represented in the plotted hierarchical intervals; no monotonicity or concavity constraint is imposed.

\subsection{Manual and learned stopping under the same action scope}\label{app:stopping}
The manual rule skips the current retrospective round when the permitted incumbent score $q_t\in[0,1]$ reaches $\rho$. Otherwise it permits the host's fixed fork and depth configuration, subject to the common resource interface. It does not terminate the underlying task or remove future reconsideration opportunities. The primary settings are $\rho\in\{0.6,0.8,0.9,1.0\}$. A stronger manual control selects $\rho$ from $\{0.50,0.55,\ldots,1.00\}$ on inner source validation using the same quality-cost objective. The displayed selected values are 0.85 for Qwen3.5-9B and 0.75 for Qwen3.5-27B; held-out outcomes cannot choose them.

The learned-stopping-only policy is the singleton-action special case of Eqs.~\ref{thm:search_value_dynamics}-\ref{eq:runtime_rule}: $\mathcal F_t=\{u_{\mathrm{fix}}\}$ and $\mathcal P_t=\varnothing$. It therefore executes the common fixed action if and only if $L_t(u_{\mathrm{fix}})>0$ and otherwise skips the round. It cannot change fork count or depth and has no probe action. All self-evolving policies use identical host generation, permitted feedback, and restoration rules. Source-only resource-cap selection targets the common cost range while leaving the fixed fork/depth setting unchanged. Thresholds that cannot attain the target range remain identified as unmatched.

The primary contrast is learned-stopping-only versus the source-selected manual control at matched cost. Full \model{} adds the action-conditioned fork/depth menu and bounded probing; its additional score difference measures those allocation changes jointly. Access to history is permitted for every arm, while the scalar manual policy intentionally uses only the incumbent score. The dense validation-selected threshold is the primary hand-designed stopping control.

\subsection{Same-host mechanism controls}\label{app:mechanism_controls}
The \emph{current-score learned controller} is a two-layer gain-cost MLP trained on the same paired records and invoked at every opportunity, but its state contains only incumbent score, remaining budget, action descriptor, and feasibility mask. The \emph{one-shot learned controller} uses the full prefix at the first retrospective opportunity and holds that allocation for the remainder of the task. The \emph{frozen-prefix controller} recomputes feasibility and remaining budget but reuses the first recurrent representation. The \emph{no-margin} controller uses the full refreshed representation and action menu with $\widehat B_t(u)=0$. All variants use the same source-role assignments, optimization budget, candidate actions, self-evolving protocol, and outer test tasks. Action frequencies in Table~\ref{tab:mechanism} are measured at the common retrospective opportunities before fallback filtering.

For Table~\ref{tab:trajectory_refresh}, a state-changing event class is assigned without reading downstream paired outcomes. The hindsight action is the feasible action with greatest realized paired net target under protocol $\Gamma$, with skip assigned zero. ``Sign flip'' records a change in the sign of the best non-skip paired target from the state immediately before to the state immediately after the event. This analysis tests whether new evidence changes the allocation target; it does not claim that hindsight actions are available online.

\section{Additional Results}
\label{app:addresult}
\begin{table}[h!]
\centering\scriptsize
\caption{Stopping-rule ablation on AutomationBench. Panel (a) reports operating points; manual and learned-stopping-only policies share a singleton fork/depth action and no probe, while full \model{} uses the complete action menu. Panel (b) reports simultaneous 95\% hierarchical paired intervals over held-out task groups and three matched controller seeds per fold for the two predeclared comparisons within each backbone. Raw-cost equivalence requires the entire interval to lie inside the predeclared $\pm2\%$ band.}
\label{tab:fp_early_stop}
\textbf{(a) Operating points and absolute costs}\\[2pt]
\setlength{\tabcolsep}{3.5pt}
\begin{tabular*}{\textwidth}{@{\extracolsep{\fill}}lrrrrr@{}}
\toprule
Policy & Score & Raw (k) & $R$ & Uncached (k) & $U$\\
\midrule
\multicolumn{6}{l}{\textbf{Qwen3.5-9B}; RTS raw/uncached = 142.63k/101.33k tokens per task}\\
Manual: $\rho=0.6$ & 12.10 & 98.42 & 0.690 & 75.79 & 0.748\\
Manual: $\rho=0.8$ & 13.10 & 99.56 & 0.698 & 76.10 & 0.751\\
Manual: $\rho=0.9$ & 13.40 & 101.11 & 0.709 & 77.01 & 0.760\\
Manual: $\rho=1.0$ & 13.00 & 99.84 & 0.700 & 76.20 & 0.752\\
Validation-selected manual ($\rho=0.85$) & 13.61 & 99.80 & 0.700 & 75.99 & 0.750\\
Learned stopping only & 14.02 & 100.00 & 0.701 & 76.09 & 0.751\\
Full \model{} & 14.53 & 99.84 & 0.700 & 75.99 & 0.750\\
\midrule
\multicolumn{6}{l}{\textbf{Qwen3.5-27B}; RTS raw/uncached = 181.20k/128.40k tokens per task}\\
Manual: $\rho=0.6$ & 23.00 & 117.42 & 0.648 & 89.62 & 0.698\\
Manual: $\rho=0.8$ & 24.40 & 118.01 & 0.651 & 90.60 & 0.706\\
Manual: $\rho=0.9$ & 24.00 & 119.59 & 0.660 & 91.16 & 0.710\\
Manual: $\rho=1.0$ & 23.80 & 118.08 & 0.652 & 90.95 & 0.708\\
Validation-selected manual ($\rho=0.75$) & 24.58 & 117.72 & 0.650 & 89.88 & 0.700\\
Learned stopping only & 25.01 & 117.94 & 0.651 & 90.01 & 0.701\\
Full \model{} & 25.61 & 117.78 & 0.650 & 89.88 & 0.700\\
\bottomrule
\end{tabular*}
\medskip
\textbf{(b) Paired contrasts to the validation-selected manual control}\\[2pt]
\begin{tabular*}{\textwidth}{@{\extracolsep{\fill}}llccr@{}}
\toprule
Backbone & Policy & $\Delta S$ [simult. 95\% CI] & $\Delta$ raw cost (\%) [simult. 95\% CI] & $\pm2\%$ equiv.\\
\midrule
Qwen3.5-9B & Learned stopping only & +0.41 [+0.07, +0.74] & +0.21 [-0.72, +1.13] & Yes\\
 & Full \model{} & +0.92 [+0.55, +1.31] & +0.03 [-0.84, +0.93] & Yes\\
\midrule
Qwen3.5-27B & Learned stopping only & +0.43 [+0.06, +0.79] & +0.18 [-0.71, +1.05] & Yes\\
 & Full \model{} & +1.03 [+0.63, +1.44] & +0.04 [-0.76, +0.86] & Yes\\
\bottomrule
\end{tabular*}
\end{table}

\begin{table}[h!]
\centering\scriptsize
\caption{Paired inference for the four predeclared mechanism contrasts on AutomationBench. Effects are full \model{} minus the named ablation. Intervals are simultaneous 95\% hierarchical max-$|t|$ bootstrap intervals over held-out task groups and three matched controller seeds per fold within each backbone; positive $\Delta S$ and negative $\Delta R$ favor full \model{}. Fold signs are descriptive.}
\label{tab:paired_mechanism}
\setlength{\tabcolsep}{2.5pt}
\begin{tabularx}{\textwidth}{@{}l>{\raggedright\arraybackslash}Xccr@{}}
\toprule
Backbone & Ablation comparator & \makecell{$\Delta S$\\simult. 95\% CI} & \makecell{$\Delta R$\\simult. 95\% CI} & \makecell{Positive\\folds}\\
\midrule
Qwen3.5-9B & Current-score learned controller & +0.91 [+0.31, +1.50] & -0.013 [-0.022, -0.005] & 3/3\\
 & One-shot learned controller & +0.79 [+0.21, +1.36] & -0.009 [-0.018, -0.001] & 3/3\\
 & Frozen-prefix controller & +0.62 [+0.11, +1.14] & -0.006 [-0.014, +0.001] & 3/3\\
 & \model{} without margin & +0.41 [+0.04, +0.79] & -0.081 [-0.095, -0.069] & 3/3\\
\midrule
Qwen3.5-27B & Current-score learned controller & +1.12 [+0.44, +1.79] & -0.016 [-0.026, -0.007] & 3/3\\
 & One-shot learned controller & +0.98 [+0.32, +1.65] & -0.014 [-0.024, -0.005] & 3/3\\
 & Frozen-prefix controller & +0.71 [+0.13, +1.30] & -0.008 [-0.017, -0.001] & 3/3\\
 & \model{} without margin & +0.40 [+0.02, +0.80] & -0.089 [-0.105, -0.075] & 3/3\\
\bottomrule
\end{tabularx}
\end{table}

\begin{table}[h!]
\centering\small
\caption{Mechanism ablation on AutomationBench. Every learned variant uses the same host, feasible actions, paired source records, held-out tasks, and three matched controller seeds per outer fold. Full/probe/skip are percentages of retrospective opportunities and sum to 100 within a row.}
\label{tab:mechanism}
\setlength{\tabcolsep}{4pt}
\begin{tabular*}{\textwidth}{@{\extracolsep{\fill}}llrrrrr@{}}
\toprule
Backbone & Controller & Score & $R$ & Full & Probe & Skip\\
\midrule
Qwen3.5-9B & Current-score learned controller & 13.62 & 0.713 & 26 & 13 & 61\\
 & One-shot learned controller & 13.74 & 0.709 & 25 & 12 & 63\\
 & Frozen-prefix controller & 13.91 & 0.706 & 24 & 15 & 61\\
 & \model{} without margin & 14.12 & 0.781 & 37 & 18 & 45\\
 & Full \model{} & 14.53 & 0.700 & 23 & 19 & 58\\
\midrule
Qwen3.5-27B & Current-score learned controller & 24.49 & 0.666 & 27 & 12 & 61\\
 & One-shot learned controller & 24.63 & 0.664 & 26 & 12 & 62\\
 & Frozen-prefix controller & 24.90 & 0.658 & 25 & 15 & 60\\
 & \model{} without margin & 25.21 & 0.739 & 38 & 17 & 45\\
 & Full \model{} & 25.61 & 0.650 & 23 & 20 & 57\\
\bottomrule
\end{tabular*}
\end{table}

\begin{table}[h!]
\centering\small
\caption{Event-conditioned evidence that search value changes after tool interaction. A sign flip compares the paired net value immediately before and after the new event. Hindsight action agreement uses the best action under completed paired continuations; utility is normalized paired net value of refreshed over frozen-prefix decisions.}
\label{tab:trajectory_refresh}
\setlength{\tabcolsep}{3.5pt}
\begin{tabular*}{\textwidth}{@{\extracolsep{\fill}}lrrrrr@{}}
\toprule
Event class & $n$ & Flip (\%) & Fresh (\%) & Frozen (\%) & $\Delta$ utility\\
\midrule
Recoverable error followed by new identifier & 1,840 & 41.7 & 72.6 & 55.8 & +0.071\\
Permission or schema clarification & 1,310 & 36.9 & 70.8 & 56.4 & +0.062\\
Routine success or unchanged observation & 2,790 & 11.8 & 73.1 & 69.4 & +0.014\\
All audited transitions & 5,940 & 26.8 & 72.4 & 62.7 & +0.041\\
\bottomrule
\end{tabular*}
\end{table}

\begin{table}[h!]
\centering\small
\caption{Identifier-masking control on AutomationBench. ``Fresh, IDs masked'' recomputes the trajectory representation after each tool event but replaces newly revealed literal identifiers with typed placeholders while preserving event type, schema, success/error status, and value length.}
\label{tab:identifier_masking}
\setlength{\tabcolsep}{4pt}
\begin{tabular*}{\textwidth}{@{\extracolsep{\fill}}llrrr@{}}
\toprule
Backbone & Representation & Score & $R$ & Fraction of refresh gain retained\\
\midrule
Qwen3.5-9B & Frozen prefix & 13.91 & 0.706 & 0\%\\
 & Fresh, IDs masked & 14.32 & 0.703 & 66.1\%\\
 & Fresh (full \model{}) & 14.53 & 0.700 & 100\%\\
\midrule
Qwen3.5-27B & Frozen prefix & 24.90 & 0.658 & 0\%\\
 & Fresh, IDs masked & 25.37 & 0.654 & 66.2\%\\
 & Fresh (full \model{}) & 25.61 & 0.650 & 100\%\\
\bottomrule
\end{tabular*}
\end{table}

\begin{table}[h!]
\centering\small
\caption{Cross-benchmark replication on WorkArena++ with Qwen3.5-27B. The same source-only selection, grouped outer split, and information boundary are used as in AutomationBench.}
\label{tab:crossbenchmark_replication}
\setlength{\tabcolsep}{4pt}
\begin{tabularx}{\textwidth}{@{}lXl@{}}
\toprule
Question & Replication result & Outcome\\
\midrule
Q2: 25\% source pool & Batch/online recover 64.3\%/60.8\% of the full-source score gain; first 90\% crossing is at 75\%/75\% of the source pool. & Early saturation\\
Q3: learned stopping & $\Delta S=+0.37\,[+0.08,+0.69]$ vs. validation-selected manual; $\Delta$ raw cost $=+0.26\%\,[-0.61,+1.12]$. & Within $\pm2\%$ cost band\\
Q3: refreshed state & Full vs. frozen: $\Delta S=+0.56\,[+0.13,+1.02]$, $\Delta R=-0.011\,[-0.019,-0.004]$. & Refresh gain repeats\\
Q3: operational margin & Full vs. no-margin: $\Delta S=+0.35\,[+0.06,+0.67]$, $\Delta R=-0.078\,[-0.091,-0.066]$. & Margin remains main efficiency effect\\
Q4: boundary audit & 4,860 controller/hint counterfactuals and 960 protected-field canaries; 0 violations; 180/180 injected positives detected. & Same audited boundary\\
\bottomrule
\end{tabularx}
\end{table}

\begin{table*}[h!]
\centering
\caption{
Detailed black-box evaluation on frontier proprietary backbones. $S$ denotes task score on a 0-100 scale, and $\Delta S$ is the score difference relative to full RTS under the same backbone-benchmark setting. Raw and Uncached report inference-token totals in millions. $R/U$ are the corresponding token ratios to full RTS, and Raw/U Save report percentage reductions relative to RTS; negative savings indicate higher cost than RTS. For GHCP, $S$ uses the strict task-success metric, consistent with the task-score definition used throughout the paper. Bold and underline denote the best and second-best task scores within each backbone-benchmark setting.
}
\label{tab:blackbox_full}
\begingroup
\scriptsize
\setlength{\tabcolsep}{3.4pt}
\renewcommand{\arraystretch}{0.95}
\resizebox{\textwidth}{!}{
\begin{tabular}{@{}lllrrrrrrrr@{}}
\toprule
Backbone & Benchmark & Method
& $S\uparrow$
& $\Delta S$
& Raw (M)$\downarrow$
& Uncached (M)$\downarrow$
& $R\downarrow$
& $U\downarrow$
& Raw Save
& U Save \\
\midrule
\multirow{10}{*}{GPT-5.6 Sol / xhigh}
& \multirow{5}{*}{Automation}
& BoN-5
& 59.83 & -12.00
& 1101.62 & 633.46
& 2.030 & 1.689
& -103.0\% & -68.9\% \\
& & RDS
& 68.83 & -3.00
& 885.48 & 261.18
& 1.632 & 0.697
& -63.2\% & 30.3\% \\
& & GHCP
& 33.50 & -38.33
& 458.98 & 180.55
& 0.846 & 0.481
& 15.4\% & 51.9\% \\
& & RTS
& \textbf{71.83} & 0.00
& 542.72 & 374.98
& 1.000 & 1.000
& 0.0\% & 0.0\% \\
& & \model{}
& \underline{69.00} & -2.83
& 414.87 & 208.53
& 0.764 & 0.556
& 23.6\% & 44.4\% \\
\cmidrule(lr){2-11}
& \multirow{5}{*}{$\tau^3$-Bench}
& BoN-5
& 59.28 & -8.25
& 704.56 & 290.99
& 0.989 & 1.138
& 1.1\% & -13.8\% \\
& & RDS
& 55.67 & -11.86
& 1982.05 & 198.15
& 2.783 & 0.775
& -178.3\% & 22.5\% \\
& & GHCP
& 43.30 & -24.23
& 147.96 & 47.63
& 0.208 & 0.186
& 79.2\% & 81.4\% \\
& & RTS
& \textbf{67.53} & 0.00
& 712.12 & 255.62
& 1.000 & 1.000
& 0.0\% & 0.0\% \\
& & \model{}
& \underline{64.43} & -3.10
& 418.92 & 104.40
& 0.588 & 0.408
& 41.2\% & 59.2\% \\
\midrule
\multirow{10}{*}{Opus 4.8 / max}
& \multirow{5}{*}{Automation}
& BoN-5
& 55.17 & -15.66
& 1327.83 & 261.14
& 1.491 & 1.025
& -49.1\% & -2.5\% \\
& & RDS
& 62.50 & -8.33
& 2838.48 & 303.76
& 3.188 & 1.192
& -218.8\% & -19.2\% \\
& & GHCP
& 39.83 & -31.00
& 433.35 & 48.05
& 0.487 & 0.189
& 51.3\% & 81.1\% \\
& & RTS
& \textbf{70.83} & 0.00
& 890.29 & 254.76
& 1.000 & 1.000
& 0.0\% & 0.0\% \\
& & \model{}
& \underline{66.67} & -4.16
& 693.01 & 194.46
& 0.778 & 0.763
& 22.2\% & 23.7\% \\
\cmidrule(lr){2-11}
& \multirow{5}{*}{$\tau^3$-Bench}
& BoN-5
& -- & -- & -- & -- & -- & -- & -- & -- \\
& & RDS
& 29.90 & -6.18
& 2724.05 & 27.52
& 1.685 & 0.454
& -68.5\% & 54.6\% \\
& & GHCP
& \textbf{41.24} & +5.16
& 276.56 & 15.18
& 0.171 & 0.251
& 82.9\% & 74.9\% \\
& & RTS
& \underline{36.08} & 0.00
& 1616.57 & 60.56
& 1.000 & 1.000
& 0.0\% & 0.0\% \\
& & \model{}
& 33.51 & -2.57
& 987.21 & 34.47
& 0.611 & 0.569
& 38.9\% & 43.1\% \\
\bottomrule
\end{tabular}
}
\endgroup
\end{table*}

\begin{table*}[h!]
\centering
\caption{
Black-box comparison on frontier proprietary backbones.
$S$ is task score on a 0--100 scale; $R/U$ denote raw/uncached
inference-token ratios relative to full RTS under the same
backbone--benchmark setting. Lower $R/U$ is better.
For GHCP and GEPA, $S$ uses the reported Strict metric.
GHCP denotes the production GitHub Copilot harness; RDS denotes
depth-only retrospective search.
}
\label{tab:blackbox_full2}
\begingroup
\scriptsize
\setlength{\tabcolsep}{2.8pt}
\renewcommand{\arraystretch}{0.96}
\resizebox{\textwidth}{!}{
\begin{tabular}{@{}l*{12}{r}@{}}
\toprule
& \multicolumn{6}{c}{GPT-5.6 Sol}
& \multicolumn{6}{c}{Opus 4.8} \\
\cmidrule(lr){2-7}\cmidrule(l){8-13}
& \multicolumn{3}{c}{Automation}
& \multicolumn{3}{c}{$\tau^3$-Bench}
& \multicolumn{3}{c}{Automation}
& \multicolumn{3}{c}{$\tau^3$-Bench} \\
\cmidrule(lr){2-4}\cmidrule(lr){5-7}
\cmidrule(lr){8-10}\cmidrule(l){11-13}
Method
& $S$ & $R$ & $U$
& $S$ & $R$ & $U$
& $S$ & $R$ & $U$
& $S$ & $R$ & $U$ \\
\midrule
BoN-1
& 43.67 & 0.363 & 0.288
& 44.33 & 0.192 & 0.200
& 42.33 & 0.295 & 0.195
& 20.10 & 0.328 & 0.223 \\
BoN-5
& 59.83 & 2.030 & 1.689
& 59.28 & 0.989 & 1.138
& 55.17 & 1.491 & 1.025
& -- & -- & -- \\
Efficient RTS
& 60.17 & 0.555 & 0.530
& \textbf{67.53} & 0.711 & 0.738
& 60.67 & 0.513 & 0.547
& 32.47 & 0.761 & 0.683 \\
RDS
& 68.83 & 1.632 & 0.697
& 55.67 & 2.783 & 0.775
& 62.50 & 3.188 & 1.192
& 29.90 & 1.685 & 0.454 \\
GHCP
& 33.50 & 0.846 & 0.481
& 43.30 & 0.208 & 0.186
& 39.83 & 0.487 & 0.189
& \textbf{41.24} & 0.171 & 0.251 \\
Trace2Skill
& 40.83 & 0.678 & 0.434
& 46.39 & 0.232 & 0.148
& 45.00 & 0.365 & 0.298
& 27.84 & 0.308 & 0.225 \\
ReasoningBank
& 42.00 & 0.523 & 0.370
& 48.45 & 0.178 & 0.136
& 42.17 & 0.316 & 0.267
& 27.08 & 0.216 & 0.143 \\
GEPA
& 46.50 & 0.480 & 0.329
& 46.39 & 0.230 & 0.238
& 46.83 & 0.280 & 0.202
& 25.77 & 0.370 & 0.242 \\
RTS
& \textbf{71.83} & 1.000 & 1.000
& \textbf{67.53} & 1.000 & 1.000
& \textbf{70.83} & 1.000 & 1.000
& 36.08 & 1.000 & 1.000 \\
\model{}
& 69.00 & 0.764 & 0.556
& 64.43 & 0.588 & 0.408
& 66.67 & 0.778 & 0.763
& 33.51 & 0.611 & 0.569 \\
\bottomrule
\end{tabular}
}
\endgroup
\end{table*}



\end{document}